\documentclass[12pt]{article}

\RequirePackage{amsmath,amssymb}

\usepackage{blurb_style}

\usepackage{url}

\usepackage{algpseudocode}
\usepackage{algorithm}

\usepackage{graphicx}
\usepackage{epstopdf}

\usepackage{fancyvrb}

\def\regC{C}

\RecustomVerbatimEnvironment{Verbatim}{Verbatim}{xleftmargin=1.5cm}

\newcommand{\inner}[2]{\left\langle #1, #2 \right\rangle}

\begin{document}
\pagenumbering{arabic}

\title{Conducting sparse feature selection on arbitrarily long phrases in text corpora with a focus on interpretability}
\author{Luke Miratrix, Robin Ackerman}
\maketitle

\begin{center}
Disclaimer: The analyses, opinions, and conclusions expressed in this paper do not necessarily reflect the views of the United States Department of Labor.

\vspace{3mm}

 Note: This paper has been accepted to \textsl{Statistical Analysis and Data Mining}. 

 Please see proofed, etc., version there.
\end{center}

\begin{abstract}
We propose a general framework for topic-specific summarization of large text corpora, and illustrate how it can be used for analysis in two quite different contexts: an OSHA database of fatality and catastrophe reports (to facilitate surveillance for patterns in circumstances leading to injury or death) and legal decisions on workers' compensation claims (to explore relevant case law).  Our summarization framework, built on sparse classification methods, 
is a compromise between simple word frequency based methods currently in wide use, and more heavyweight, model-intensive methods such as Latent Dirichlet Allocation (LDA).  
For a particular topic of interest (e.g., mental health disability, or carbon monoxide exposure), we regress a labeling of documents onto the high-dimensional counts of all the other words and phrases in the documents.  
The resulting small set of phrases found as predictive are then harvested as the summary.  Using a branch-and-bound approach, this method can be extended to allow for phrases of arbitrary length, which allows for potentially rich summarization.  We discuss how focus on the purpose of the summaries can inform choices of tuning parameters and model constraints.
We evaluate this tool by comparing computational time and summary statistics of the resulting word lists to three other methods in the literature.
We also present a new R package, \verb|textreg|. 
Overall, we argue that sparse methods have much to offer text analysis, and is a  branch of research that should be considered further in this context.  
\end{abstract}

\section{Introduction}

Regularized high dimensional regression can extract meaningful information from large text corpora by producing key phrase summaries that capture how a specific set of documents of interest differ from some baseline collection.
This text summarization approach has been called Concise Comparative Summarization (CCS) \cite{Jia:2014ko}, underscoring two fundamental features of this tool: (1) the comparison of a class of documents to a baseline or complete set in order to remove generic terminology and characteristics of the overall corpus; and (2) the resulting production of a short, easy-to-read summary comprised of key phrases.
Such summaries can be useful for understanding what makes a document collection distinct and can be used to inform media analysis,  understand incident reports, or investigate trends in legal decisions.

Many classic methods of text summarization tend to focus on single words or short phrases only.
Approaches such as Latent Dirichlet Allocation \cite{blei2003latent} also do not extend naturally to phrases.
On the other hand, one regression-based method \cite{Ifrim:2008uka,Ifrim:2010vf} that does allow for longer phrases does not allow for rescaling of the counts of phrases in the text based on the overall frequency of appearance of such phrases, which can negatively impact the quality of resulting summaries.
In this paper we merge two CCS approaches to allow for rescaled arbitrary-length key phrases that can include gaps.
We briefly discuss how this is done below.
Our new CCS tool be easily used via our new R package, \verb|textreg|, which allows for rapid exploration of text corpora of up to a few gigabytes in size.

Even given these tools, when a researcher desires to conduct a specific analysis, he or she is faced with many choices. 
In particular, the implementation and regularization of the regression itself can be done in several ways---and the impact of choosing among these ways is one of the foci of this paper.
In particular, we argue that if the researcher has specific goals for interpretation in mind, these goals can inform choice of tuning parameters.
For example, when faced with a corpus where only a few documents are of interest and the rest are to be used as a baseline, a researcher may choose to allow only positive weights on phrases, in order to simplify interpretation.  
Similarly, choice of tuning parameter can be governed by a researcher's level of interest in pruning rare phrases.
We also offer a method for testing for a significant relationship between the text and the labeling that also provides a threshold regularization value.


We compare this tool to other related state-of-the-art methods. First, we compare to multinomial inverse regression (MNIR) \cite{Taddy:2013tp}, a text regression method that is primarily designed to be distributed across many cores in order to be able to handle massive data.
We also compare to a classic linear Lasso approach (see, e.g., \cite{Tibshirani:2004tz}), which is similar to this method run on pre-computed document-term matrices without some of the flexibility.
We finally compare to the original Ifrim et al. method that is one of the building blocks of this work.
In these comparisons we investigate computation time, prediction accuracy, and different features of the resulting word lists.
The different approaches give very different types of lists, and we hope this work gives some guidance to the practitioner as to how to sort through the options.
 
As a case study we use this tool to examine a large collection of occupational fatality and catastrophe reports generated by the Occupational Safety and Health Administration (OSHA) in the United States. 
As a motivating example, we examine hazardous exposure to methylene chloride, a neurotoxin, during bathtub refinishing operations. 
In 2013, OSHA and the National Institute for Occupational Safety and Health (NIOSH) jointly issued a Hazard Alert calling attention to a recurring pattern of this nature following the deaths of least 14 workers since 2000 in related circumstances. However, the sheer volume of information describing occupational fatalities and catastrophes may have initially obscured this pattern in the years preceding its detection. Although OSHA maintains a database of narrative reports describing fatalities and catastrophes (``Fat/Cats''), similar patterns of preventable exposure to occupational hazards may be difficult to identify efficiently through manual review alone, given the large number of narratives in this database. 
Thus, using methylene chloride as a case study, we consider whether text mining techniques can help identify important patterns in circumstances of hazardous exposures.

In our framework, a summary list of key words and phrases ideally represents and reveals the overall content of a collection of narrative reports. 
For example, one summary for all narratives related to ``Methylene Chloride'' contained the words ``bathtub'' and ``stripper.''
To qualitatively evaluate our tool, we manually examine these words and phrases in the context of the original reports and consider whether our text summarization tool effectively characterizes the circumstances of the bathroom refinishing fatalities.  
 

In general, we explore whether we can construct text mining algorithms that, when applied to an entire corpus, can uncover ``needles in the haystack'' patterns such as the connection between bathroom refinishing and overexposure to methylene chloride.
At this stage we are not focused on rates or relative risks of particular patterns; we are instead focused on the crude detection of textual patterns that may represent meaningful information about how certain types of injuries and fatalities occur. Such findings, even if they involve only a few recorded deaths or injuries, may facilitate the prevention of many future fatalities, particularly in the context of emerging hazards. 

 We also examine our tool's ability to extract information from a collection of legal decisions from the Employees' Compensation Appeals Board (ECAB), which handles appeals of determinations of the Office of Workers' Compensation Programs (OWCP) in the U.S. Department of Labor (DOL). Here we investigate what information we can extract about different categories of cases. In particular, we examine cases involving a question as to whether the work environment caused a mental health condition (an ``emotional condition,'' in the parlance of ECAB). 
We find that while the CCS tool does extract meaningful information relating to the cases of interest, further work needs to be done to obtain more nuanced summaries.

Overall, the CCS approach does allow for exploration of text and does extract meaningful information.
Extending earlier, fixed-length phrase tools to allow for longer phrases and phrases with gaps does increase the expressive capability of the summaries.
The methods for picking a tuning parameter, while possibly a bit aggressive and conservative, do provide an alternative paradigm for data analysis with an eye to extracting human meaning from text.

\section{Overview of Summarization}

This paper extends the concept of Concise Comparative Summarization (CCS) discussed in \cite{Jia:2014ko}, incorporating a prior approach proposed by \cite{Ifrim:2008uka} to result in an overall improved methodology.
Concise Comparative Summarization involves comparing a pre-specified set of documents to a baseline set. One can think of it as a regularized regression of some labeling of the $m$ documents (normally +1 and -1) onto the collection of all possible summary key phrases.
For example, in one analysis we label documents relating to incidents involving carbon monoxide (CO) as $+1$ and the remaining documents as $-1$.
Each potential key phrase, or ``feature,'' is considered to be a covariate in the regression, and is in principle represented as an $m$-vector of measures of the feature's presence (e.g., appearance count) in each of the $m$ documents.
By using sparse regularization, only a small subset of these possible summary key phrases is selected. These phrases are taken as the final summary.
So, for example, the resulting phrases of our CO-related regression would ideally indicate what is different about CO-related events when compared to other workplace injuries and fatalities.
At root we are taking those phrases most useful for classifying a set of documents by the given labeling as the summary of the positive set of documents as compared to the negative, baseline, set.

It is worth emphasizing that the focus is not predictive quality; the selected features themselves are the object of interest and the quality can only be measured by their usefulness in the original human interpretation-based question that motivated the exploration.
Thus, these methods in principle require human analog experiments to validate their findings.
This can be done; see \cite{Grimmer:2010hp,Chang:2009wd}, or \cite{Jia:2014ko} for examples.
We are using text classification tools, but the classification is a byproduct.
There are many possible choices for how to implement this regression including whether to use logistic or linear regression, and whether to rescale the frequency appearance of the phrases before regression. Prior work has shown that rescaling phrase frequency is quite important; failing to appropriately do so can result in summaries that have little informative content even while predictive accuracy is maintained.
This is not surprising; term frequency in text is a known and serious concern when data mining large text corpora (e.g., \cite{Jia:2014ko, Bischof:2012ud}), as was first illustrated in the information retrieval literature (e.g., \cite{Salton:1988wx,salton1991dat}).

In text classification, and, by extension, key-phrase extraction via text classifiers, there is some desire to allow for phrases as well as unigram (single word) features.  
One approach is to calculate and use all phrases up to $n$ words long as the overall feature set.  
For long phrases this can quickly become intractable as there is a blowup in the number of possible phrases a corpus may contain.
To solve this problem, Ifrim et al. \cite{Ifrim:2010vf,Ifrim:2008uka} allow for arbitrary length phrases by generating the features ``on-the-fly'' as part of the optimization.
As an added benefit, this approach easily allows for ``gaps,''  i.e. phrases with wildcard words, which greatly enhances the potential expressiveness of the resultant summaries.

Ifrim et al.'s algorithm, based on work of \cite{Kudo:2004wa} and, even earlier, \cite{Schapire:2000us}, fits an elastic-net-penalized logistic regression with the features consisting of the entire space of all phrases.
(Also see \cite{genkin07} or \cite{Zhang:2001wx} for other examples of regression on text and \cite{Hastie:2003wp} or \cite{Zou:2005wy} for an overview of elastic nets and other regularized regression methods in general.)
Ifrim et al. initially propose an algorithm to solve a penalized logistic regression of:
\[ \hat{\beta} = \arg \min_{\beta = (\beta_1, \ldots, \beta_p)} \sum_{i=1}^m y_i c_i'\beta + \log \left( 1 + \exp( y_i c_i'\beta ) \right ) + \regC R( \beta ) \]
with $c_i$ being the feature vector for document $i$ with the $c_{ij}$ as binary indicators of the presence of feature $j$ in document $i$, $y_i$ being the $-1/1$ class label, $p$ being the number of features including all phrases, $C$ being a regularization tuning parameter, and $R(\beta)$ being some regularization function.
They later extend this to allow for alternate loss functions such as a hinge loss.
However, they do not allow for rescaling features.
By modifying their methods, we show how rescaling can be incorporated into their overall approach. 
They also do not allow for an intercept term, which can introduce difficulty with the summarization process if the number of positive features is not close to 50\%.
We extend their algorithm to allow for a (non-penalized) intercept term as discussed in \cite{Jia:2014ko}.
We implement these modifications by extending their code, and then wrap the resulting algorithm in a new R  package, \verb|textreg|, to make it easier to use in a rapid and exploratory manner.
We also provide some useful tools and visualizations to help researchers understand the resulting summaries.
  
The core idea behind the algorithm is a greedy coordinate descent coupled with a branch-and-bound algorithm.  With each step, the algorithm searches the entire feature space for the feature that has the highest gradient at the current $\beta$.  
This is obviously a very large search space, but it can be pruned using a relationship that bounds the size of a gradient of a sub-phrase by a known calculation on any parent phrase.  
In the search we track the current optimal feature, and then for each new feature considered, if the bound on all the children of that feature is smaller than the current optimum, prune all those children from the search. 

\subsection{Related work}

CCS is distinct from classification.  Classification is focused on sorting documents, such as for attributing authorship \cite{Mosteller:1984vx,Airoldi:2006vs} or counting types of news events \cite{King:2003toa,Hopkins:2010tua}. 
Text classification has been attempted using a wide array of machine learning methods such as naive bayes, linear discriminant analysis, or support vector machines \cite{Cortes:1995wa}, which easily allow for large numbers of features (the words and phrases) to be incorporated.  
For comparisons of these different methods on text see \cite{foreman2003bns}, \cite{Airoldi:2006vs} or \cite{Dumais:1998to}.
For SVMs as an approach in particular for text, see the book of \cite{joachims2002}.
For such methods and these evaluations, however, the features themselves are not of primary interest, classification is.  
We instead attempt to extract meaning from documents by contrasting sets to each other.
This is most similar to key phrase extraction, a literature in its own right.  
See, for example, \cite{roseetal2010}, \cite{Frank:1999vt}, or \cite{Chen:2006wba}.

Interpreting text is a difficult task, and can be done in a variety of ways.  
For example, \cite{Gerrish:2011wy} use text to predict roll call votes with an LDA (Latent Dirichlet Allocation) algorithm \cite{blei2003latent} in order to understand how language of law is correlated with political support.  \cite{Grimmer:2010hp} model political text to explore who dominates policy debates.  
Truly validating a finding is generally quite difficult and requires a great deal of effort to do properly, as these papers illustrate quite well.
 Our tools are primarily intended for exploration; validation is not within CCS's scope without additional human validation efforts or  alternative techniques.

Of the many approaches to text analysis, variations of LDA \cite{blei2003latent} in particular have recently been widely investigated and used.
These consist of a Bayesian hierarchical model that describe documents in a corpus as a mixture of some number of different distributions on the words.
They can reveal structure in a corpus and have been shown to capture human meaning.
However, generating novel models for specific circumstances is difficult.
Even mild changes to the model can be technically quite challenging  and consist of an entire research topic in its own right.
They are either computationally expensive or only solved approximately (via, e.g., variational methods).
In the spirit of diversity in research approaches, we take a different path.

This is not to say that using sparse regression methods on text is new; see for example \cite{foreman2003bns}, \cite{lee2006nmt} and \cite{Yang:1997wf}.
\cite{Eisenstein:2013ws} use sparsity to model topics by representing them as small collections of  phrases that stand out against a background.  \cite{ElGhaoui:2011uo} showcase several methods, such as sparse PCA, to investigate large corpora.  There are many others.

One aspect of our approach that we believe is novel is allowing for complex features in the form of overlapping phrases, especially phrases with wildcard words, while maintaining the ability to rescale features. 
This allows great flexibility in the expressiveness of the possible summaries generated, and it is not obvious how to naturally extend methods such as LDA, which rely on a generative model where words are picked i.i.d. from some distribution, to do this.

\section{Rescaled n-gram regression}

Initial methods regress the $y_i$ on the $c_i$, where $c_i$ is either the vector of counts with $c_{ij}$ being how often phrase $j$ appears in document $i$ or of binary indicators of appearance, with the elements $c_{ij} \in \{0, 1\}$ indicating the presence of phrase $j$ in document $i$.  
This can be problematic in that common phrases (e.g., ``the,'' or, less obviously, ``usually,'') end up having much higher variance than less common ones and thus it is easier to pick them up due to random fluctuations.  
See Section~\ref{sec:rescaling} for further discussion.

Rescaling the features can correct this as pointed out in, e.g., \cite{Monroe:2008ey}.
In particular, $L^q$-rescaling (for $q \geq 1$) transforms the vectors $c_i$ into new covariate vectors $x_i$ as
\[ x_{ij} = \frac{c_{ij}}{z_j},  \mbox{ where } z_j \equiv \left( \sum_{i=1}^{n} c_{ij}^q \right)^{1/q} . \]
This is similar to standardizing columns in a Lasso regression; if you do not, then phrases with high variability need smaller coefficients to have similar impacts on prediction.
This makes them ``cheaper'' under the regularization and therefore appear more frequently due to random chance.

Once our feature space has been standardized with each phrase having an $L^q$-length of 1 we regress our $y$ onto these rescaled $x$ and an intercept.
This is a high-dimensional problem with $p \gg m$.  
As we want a small number of phrases, we use a sparse regularization $L^1$ penalty.
We also use a squared hinge loss to obtain:
\begin{align*}
 \mathcal{L}(\beta) &= \sum_{i=1}^n \left[ \left( 1 - y_i (\beta_0 + x_i'\beta) \right) \vee 0 \right]^2 + \regC  \sum_{j=1}^p |\beta_j|  ,
\end{align*}
with $a \vee b$ denoting the maximum of $a$ and $b$.  
For this loss function an \emph{over}-prediction is not penalized.
From a prediction standpoint, we wish $\beta_0 + x_i'\beta = y_i$;  
if we fall short, we have quadratic loss, if it does, the loss is zero, and if we overshoot we still have zero loss.
There is no penalty for ``over-predicting'' a document's label.
We use squared hinge loss as this is similar to the Lasso, shown to be effective in \cite{Jia:2014ko}, but also monotonic, which is needed for the optimization algorithm.
Also note the penalty term does not include the intercept, $\beta_0$.

To generalize this framework, taking notation from Ifrim et al., let our loss for an individual document be $\xi( m_i )$ with $m_i = y_i( \mu + x_i ' \beta )$.  
We can use any monotonic loss functions with $\xi' \leq 0$ everywhere.
The squared hinge loss from above is $\xi( m ) = \left(\left(1 - m\right) \vee 0\right)^2$; this is very similar to an OLS-type penalty of $(1-m)^2$.
Logistic would be $\xi( m ) = \log(1 + e^{-m})$.

Regardless of the choice of $\xi$, the loss term can be expressed in the original counts as
\begin{align*}
 \mathcal{L}(\beta) &= \sum_{i=1}^n \xi\left( 1 - y_i \left(\beta_0 + \sum_{j=1}^p \frac{c_{ij}}{z_j} \beta_j \right) \right) + \regC \sum_{j=1}^p |\beta_j| \\
		 &= \sum_{i=1}^n \xi( m_i ) + \regC \sum_{j=1}^p |\beta_j| .
\end{align*}
We then obtain $\hat{\beta}$ as
\begin{equation}
 \hat{\beta} = \arg \min_\beta \mathcal{L}(\beta)  \label{eq:main_optimization}	
\end{equation}
Alternatively, by letting $\tilde{\beta}_j \equiv \beta_j / z_j$, we can move the $z_j$ to the penalty term and regress on the counts $c_i$; this gives the identical loss as seen by considering $\tilde{m}_i = y_i( \mu + c_i ' \tilde{\beta} )$ on the rescaled columns: scaling a column by $z_j$ is the same as penalizing the associated $\beta_j$ by $z_j$.
This has ties to weighted Lasso approaches such as the adaptive Lasso \cite{zou2006adaptive}, in that we now have feature-specific penalties.
The gradients change, however, which can affect the optimization.  See Appendix B.

Solving Equation~\ref{eq:main_optimization} is done with greedy coordinate descent.  See Algorithm~\ref{alg:greedy}. 
For greedy coordinate descent, we repeatedly find the feature with the highest gradient, and then optimize its corresponding $\beta_j$ with a line search over the loss function.
Because this is a convex problem, this will converge on the global maximum as each iteration will decrease $\mathcal{L}(\beta)$ and since the gradient along all the coordinates can only be 0 if we are at a maximum.  For a proof see, e.g., \cite{Wu:2008vf} or \cite{Luo:1992wc}.
We keep a cache of all the non-zero features in our model; we do not need to ever calculate or store all possible features.
The main computational cost of the algorithm is in finding the feature with the largest gradient.
To do this, we dynamically generate the features by exploiting the nested structure of any multiword phrase having a smaller phrase as a prefix.  
This inner algorithm is shown on Algorithm~\ref{alg:findBest}.
Here we first examine all unigrams, then bigrams, and so forth until there are no more eligible phrases.  We first calculate the gradient for all unigrams and enter them into the queue.  Phrases in the queue are placeholders for their family of superphrases.  When we pull a phrase out of the queue, we check to see if we can prune all of its children and if we cannot, we determine the phrases' children, calculate gradients for these children, and finally enter them into the queue.
This algorithm would work without pruning, but if we were able to prune all the at-zero children of a feature before examining them, we could achieve large speed-ups.  
And indeed some pruning is possible due to a trick of bounding a child's gradient based on the structure of its parent, although the rescaling makes keeping this bound tight more difficult than in the original Ifrim et al. presentation. 
The main idea is if a bound on the gradients of a family of features is less in magnitude than our current best gradient, we can prune them all.
We discuss finding such bounds next.

\begin{algorithm}
  \caption{Greedy Coordinate Descent}
  \label{alg:greedy}
  \begin{algorithmic}
    \State $\beta = \emptyset$ 
	\State $features = \emptyset$
	
    \While  {Not Converged}
    \State $\beta[intercept]$ = updateFeature(intercept)
	\State $f$ = findHighestGradient
	\State $features.add( f )$
	\State $\beta[f]$ = updateFeature(f)
	\EndWhile
\end{algorithmic}
\end{algorithm}

\begin{algorithm}
  \caption{findHighestGradient}
    \label{alg:findBest}
  	\begin{algorithmic}

  \State $features$ = all non-zero features so far.
     \State $best_f = \arg \max_{f \in features} gradient( f )$
	\State $u_1, \ldots, u_{p_1}$ = all unigrams in dictionary
	\State Q = queue( ), a queue of all features to check
		\For {$u \in u_1, \ldots, u_{p_1}$}
				\If{ $\mbox{gradient}(u) > \mbox{gradient}(best_f)$ }
					\State $best_f = u$
				\EndIf
				\State Q.add( $u$ )
		\EndFor

	\While {Q is not empty}
		\State f = Q.next()
		\If {not canPrune($f, best_f$)}
			\For {$c \in \mbox{children}(f)$}
				\If{ $\mbox{gradient}(c) > \mbox{gradient}(best_f)$ }
					\State $best_f = c$
				\EndIf
				\State Q.add( $c$ )
			\EndFor
		\EndIf
	\EndWhile
	\end{algorithmic}
\end{algorithm}

\subsection{Bounding gradients}
\label{sec:bound_gradients}
Take any feature $j$ with corresponding appearance pattern across the documents $c_j$.  
For any feature $k$ with feature $j$ as a prefix, we know that $c_{ki} \leq c_{ji}$ for $i = 1, \ldots, n$, which we write as $c_k \preceq c_j$.
We also know that $c_{ki} \geq 0$ for $i = 1, \ldots, n$ because they are counts, so $0 \preceq c_k$.
I.e., given a phrase $j$, any phrase with phrase $j$ as a prefix can only have a count vector bounded between 0 and phrase $j$'s count vector.

During our search we consider phrases from shorter to longer.
For each phrase $j$ we, based on that phrase's appearance pattern in the text, calculate a bound $b_j$ on the magnitude of the highest gradient a ``best case'' hypothetical superphrase with that prefix could have.
If this $b_j$ is smaller than the current best achieved gradient $|\Lambda|$, then we can prune from consideration all phrases with phrase $j$ as a prefix because if $b_j \leq |\Lambda|$ we have for any superphrase $k$ of $j$
\[ \left| \frac{d}{d\beta_k}L(\beta) \right| \leq b_j \leq |\Lambda| . \]
Therefore we want $b_j$ to be as small as possible, i.e., tight, to make phrases easier to prune.

As derived in Appendix B, one such overall bound is, for any $q \geq 1$,
\[ \left|  \max_{k : 0 \preceq x_k \preceq x_j} \frac{d}{d \beta_k} L( \beta )  \right| \leq 0 \bigvee \max \left\{  \left( \sum_{y_i = -1, c_{ij} > 0} \left| \xi'( m_i )\right|^r \right)^{1/r},   \left( \sum_{y_i = 1, c_{ij} > 0} \left|\xi'( m_i )\right|^r \right)^{1/r} \right\} - \regC,  \]
where $1/r = 1 - 1/q$.
For any phrase $j$, these bounds can be computed by summing over only those documents that have phrase $j$, rendering them computationally tractable.  
Because the rescaling allows for theoretically very predictive yet relatively rare phrases, these bounds are unfortunately quite loose, making it hard to substantially trim the search tree.
One possible avenue for improvement would be to integrate the preprocessing step suggested by \cite{elghaoui11}.

As special cases, $q=1$ gives $r = \infty$ and a bound of the maximum $\left|\xi'(m_i)\right|$ for any document $i$ that has phrase $j$, and $q=\infty$ gives $r = 1$ and a bound of the maximum of the two sums of the $\left|\xi'(m_i)\right|$ across the negative and positive documents containing phrase $j$---which is Ifrim's bound, corresponding to no rescaling, up to the scaling of the maximum occurrence of the phrase in any single document.

All of the above is easily extended to an elastic net \cite{Zou:2005wy}.  See note in Appendix B.

\section{Choices of rescaling and additional constraints}
\label{sec:rescaling}

Choices of rescaling (e.g., the $q$ in the $L^q$-rescaling) and further restrictions on the optimization problem can focus the CCS tool on different aspects of the summary.
We can seek to generate summaries with more general or more specific words, for example, or enforce a contrast of a target set to a larger background set which eases interpretability.
We discuss how in the following subsections.

\subsection{Rescaling}
Phrases vary greatly in their overall appearance in text, with a very long tail of words and phrases that appear in nearly every document, and the bulk of phrases appearing only one or two times.
A phrase's rate of appearance is connected to its underlying variance if we represent the phrase with its count vector.
This can cause problems when selecting the most meaningful phrases.
In particular, common phrases can easily dominate because they have greater variance.
Typically this is handled with stop-word lists, which are lists of words that are a priori deemed low-information and dropped before analysis.
For a thorough discussion of this, see \cite{Monroe:2008ey}.  
And, as \cite{Monroe:2008ey} discusses, stop-word removal is finicky, not general, and imperfect.  
They can not easily be adapted to differing contexts or languages.
Furthermore, how to implement stop word removal when phrases are the object of interest is unclear.
Rescaling, however, can not only serve the function of a stop-word list but do a superior job \cite{Jia:2014ko}.

Rescaling is critical, as is widely known in information retrieval.
Without rescaling, stop words are easily selected by virtually all text mining procedures.
Even with stop words being dropped, typically the ``runner up'' most common phrases are then selected, primarily due to random variation in their appearance pattern. 
To see this, test nearly any off-the-shelf text mining tool without removing stop words first; more often than not, these methods will fail, and their results will be dominated by these low-information words.  
Stop word lists are a hard-threshold solution when a soft-threshold tapering is more appropriate.
Rescaling offers such a tapering approach.
The question then becomes which rescaling to use, or alternatively, how much tapering do we want.
We use $L^q$-rescaling because it offers a class of choices (and integrates well into the gradient descent algorithm).
With $L^q$-rescaling, different choices of $q$ weight phrases relatively differently, allowing for focus on more common or uncommon phrases at the desire of the researcher.

\begin{figure}[htbp]
\begin{center}
\includegraphics[width=0.5\textwidth]{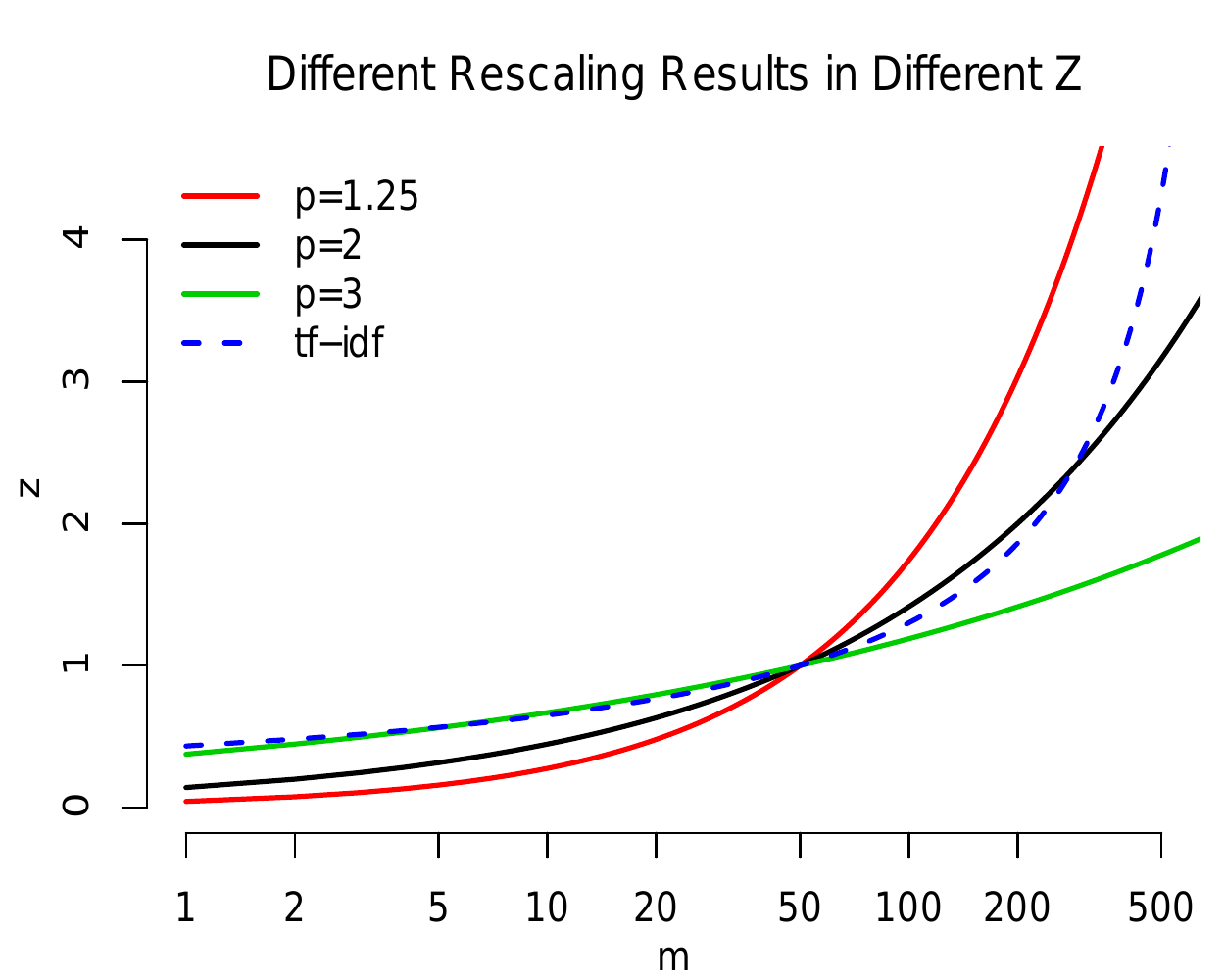}
\caption[Impact of Different Choices of Rescaling]{Impact of Different Choices of Rescaling.  Here we see different rescaling factors $z$ for phrases with a single appearance for each of $m$ documents out of 1000 total documents.  More common phrases are penalized more greatly relative to rare phrases.}
\label{fig:rescaling}
\end{center}
\end{figure}

Overall, lower $q$ means generally higher normalization factors $Z$, which will change the appropriate $\regC$ for regularization.
The main point of concern, however, is the relative sizes of the weights for rare phrases compared to common phrases.
In general, a $L^{1.25}$-rescaling heavily penalizes common phrases while a $L^3$-rescaling does not.
On the other hand, $L^3$-rescaling penalizes rare phrases slightly more than lower choices of $q$.
To illustrate, see Figure~\ref{fig:rescaling}; here we consider a sequence of phrases that appear once in each of $m$ out of 1000 documents.  
The different series of weights have been rescaled so a phrase which appears 5\% of the time (50 times) has the same weight for all choices.

\paragraph{tf-idf rescaling.}  
A related strategy for rescaling is tf-idf rescaling, from information retrieval \cite{Salton:1988wx}.  It is typically something like (variations exist):
\[ x_{ij} = \frac{c_{ij}}{n_i} \cdot r_j \qquad \mbox{with} \qquad r_j \equiv \log \frac{ n }{ d_j }, \]
with $d_j$ being the total number of documents containing the phrase $j$.
It differs in that it corrects for each document length with the $n_i$; we do not do so.  If documents are roughly the same length, this becomes less relevant.
Tf-idf also puts more of an extreme difference between weights for rare and common phrases, scaled by the total number of documents.
For rare and mid-range phrases, the tf-idf rescaling is similar to a $L^q$-rescaling with large $q$.

\subsection{Interpretation and negative coefficients}

CCS returns a list of phrases with non-zero coefficients $\beta$.  
Interpreting these coefficients can be quite difficult.
Just as in OLS, a model-based interpretation for $\beta_k$ would be that changing feature $k$ by 1 would change the prediction by $\beta_k$  \emph{holding other features fixed} (which, given normalization, means changing feature $k$ by a count of 1 would change the outcome by $\beta_k / z_k$).
However, given the lack of a well-motivated model in our context, interpreting the magnitude of these coefficients is somewhat dubious.

Nonetheless, we still wish to interpret the sign of the coefficients: positive indicates a feature is associated with our positive class of documents, and negative indicates the negative class.
When the negative group is a baseline, however, it is not an object of direct interest. 
This is especially the case if it is much larger and more diverse than the positive class.
In this case the regression is ideally subtracting out baseline characteristics, leaving the researcher with what makes the positive class distinct and noteworthy.  
Here, interpreting negative coefficients can be difficult.
One interpretation would be that such features are ``conspicuous in their absence.''  
 
Unfortunately, even when there is a mix of positive and negative features, we can still end up with unclear interpretations of the sign due to the ``holding other features constant'' aspect of the above. 
For example, a negative weight for a feature might be offsetting the positive weight for a highly correlated alternate feature, and in fact both features may have a positive correlation with the labeling.
In this case, interpreting the negative sign as, e.g., conspicuous in its absence, is erroneous.
It is more accurate to say the feature is conspicuous in its absence given its normal association with the second feature.
This can be hard to communicate.
 
One solution is to extend the optimization to consider only the set of positive $\beta$, forcing negative coefficients to not exist.
This is an easy extension of the above algorithm: simply drop the lower bound on the gradient search and truncate any line-search update of a $\beta_k$ at 0.

This is not to say that negative features are useless.
For example, if we allow negative features and find all the coefficients are positive, it would suggest that the positive group has a clearer signal than the negative group.
Only phrases found in the positive group are differentiating the groups.
This might suggest distinct language use, larger vocabulary, or specific turns of phrase on the part of the positive group, which could be of interest in its own right.

\section{Picking the regularization parameter}
For most regularized regression settings, picking the regularization parameter $\regC$ is a notoriously difficult problem.  
In general, higher $\regC$s lead to fewer features, i.e., more concise summaries.  Low $\regC$ summaries will be more verbose.
However, an overly low $\regC$ allows for over-fitting, which in our context means obtaining features that are detected soley due to random fluctuations in the appearance patterns of phrases.
We need to ensure that $\regC$ is sufficiently large to mostly prune out such noise.

Classically, selection of $\regC$ is done using methods such as cross-validation to optimize prediction accuracy on out-of-sample predictions.
As prediction is not our primary focus, we look for other methods to select $\regC$ that enhance the quality or interpretability of the summaries generated.
The lack of appropriateness of prediction accuracy is somewhat motivated by the literature; prediction accuracy is, for example, not the same as model selection, as is illustrated by the choice between AIC and BIC selection methods in regression. 
We present two methods, rooted in the goals of CCS, to select $\regC$. 
The first is to conduct a permutation test to select a $\regC$ that gives a statistically significant summary in that the summary being non-empty indicates the presence of systematic differences in the text between the positively and negatively marked documents.
The second is to select a minimum $\regC$ to guarantee the pruning of very rare phrases.
We discuss how one might select which approach to use in the discussion after the case studies, below.

\subsection{A permutation test on the summary}
One might wonder if the phrases returned by CCS are simply due to random chance.
There are so many different phrases, it is reasonable to believe some will randomly be associated with any document labeling.
We can control this with a permutation test.
This is an exact test, and the resulting $p$-value is easy to interpret.

To test whether it is meaningful to generate a summary at all, repeatedly randomize the labeling across the documents, regress, and find the corresponding $\regC^*$ that zeros out all the features, given our random permutation of the labels. 
This gives a ``null distribution'' of what $\regC$ is appropriate if there were no signal.
Finally,  compare our originally observed $\regC^{obs}$ to this distribution of fake $\regC^*$s.
We calculate a $p$-value of
\[ p = \Pr \left\{ \regC^{obs} \geq \regC^* \right\}.\]

If $p$ is small, we conclude that the needed regularization to zero out our originally observed signal is greater than that for a random signal; i.e., there is a relationship between the labeling and the text.
Similarly, if we pick a $\regC$ that is at the 95\textsuperscript{th} percentile of our permutation distribution, we are 95\% confident that the resulting summary being non-empty is due to the relationship of the labeling and the text, and not due to random chance.

The individual phrases, however, are \emph{not} specifically tested as being significant; it is possible that they would change, for example, given mild perturbations to the data.
Nevertheless, this test provides a useful minimum bound for the final $\regC$.
Any $\regC$ lower than this bound could result in a non-empty summary purely due to random chance.

In a similar manner, one can check the coherence of final summaries by generating summaries under different permutations of the labeling  (potentially adjusting $\regC$ with each iteration to get similar length summaries for all permutations) and creating a list of lists with the actual summary randomly inserted. 
If humans can then reliably pick out the actual summary from the fake ones, this is indication that the structure of the summary is not due to random chance.
This idea is based on assessing the quality of EDA visualizations; see \cite{Wickham:2010vr}.

\subsection{Pruning rare phrases}
\label{subsec:prune}

Most potential phrases are rare, showing up only a handful of times in even very large corpora.  
Selecting from such phrases introduces a severe multiple testing problem, and we seek to appropriately regularize the regression with $\regC$ to solve it.
In particular, rare phrases that show up only a few times can be selected if they happen to fall only in the positive set.
More generally, with improper rescaling of features, a term that shows up once in the positive examples with a high count and several times in the negative examples with a low count can also be selected.  
This often is contrary to the interpretive goal behind selecting predictors.
We want phrases that are general summaries, informing the researcher of aspects across multiple documents.
These problems can be partially remedied by proper selection of the tuning parameter $\regC$.
Here we investigate minimal $\regC$ to guarantee that quite rare phrases are dropped.

We find such $\regC$ by investigating so-called ``perfect predictors.''  
Consider a feature $j$ such that $c_{ij} = 0$ if $y_i = -1$ and $c_{ij} = 1$ for some of the documents where $y_i = +1$.  
For the moment, assume we do not have multiple counts in any document.
This is a perfect predictor, predicting $r \equiv \sum_i c_{ij} \leq s$ of the $s$ positive documents.  
These perfect predictors could be used to identify a subset of the positive examples while incurring no loss for the negative examples.  
The only cost of including such a predictor is due to the regularization term $\regC$.
If we set $\regC$ high enough, the cost will be prohibitive and we will not select.  
In fact, the cut-off of
\[ \regC^* = 2(1-\mu)r^{1 - 1/q}, \]
with $\mu = \frac{1}{m} \hat{Y}_i$, with $\hat{Y}_i$ being the prediction for document $i$ without any such hypothetical phrase, suffices.  See Appendix B for a derivation.

For this $\regC^*$, any perfect predictor of $r$ documents will be pruned.
For comparison, see Table~\ref{tab:need_cut}, which shows (for both $\mu \approx -1$, the case of few positive and many negative examples, and $\mu \approx 0$, the case of roughly equal numbers of positive and negative documents) the needed cut-offs for  $r=1, 2, 4$.  
This cutoff will generally be overly aggressive; if other predictors also predict these documents, then the gain of including the perfect predictor is potentially less.
 
\begin{table}[ht]
\centering
\begin{tabular}{ccrrr|rrr}
\hline
  &                   &   \multicolumn{3}{c}{$\mu \approx -1$} & \multicolumn{3}{c}{$\mu \approx 0$} \\
q & $\regC^*$ & $r=1$ & $r=2$ & $r=4$ & $r=1$ & $r=2$ & $r=4$ \\
\hline
$\infty$ & $2(1-\mu)r$ & 4 & 8 & 16  & 2 & 4 & 8 \\
4 & $2(1-\mu)r^{3/4}$ & 4 & $\approx6.7$ & $\approx11.3$ & 2 & $\approx3.4$ & $\approx5.6$ \\
2 & $2(1-\mu)\sqrt{r}$ & 4 &  $4\sqrt{2}$ & 8 & 2 & $\sqrt{2}$ & 4 \\
4/3 & $2(1-\mu)r^{1/4}$ & 4 &  $\approx4.75$ & $\approx5.7$ & 2 & $\approx2.4$ & $\approx2.8$ \\
1 & $2(1-\mu)$ & 4 & 4 & 4 & 2 & 2 & 2 \\
\hline
\end{tabular}
\caption{Needed $\regC$ to Prune Rare Perfect Predictors.}
\label{tab:need_cut}
\end{table}

\paragraph{No rescaling.}
 As discussed at the end of Section~\ref{sec:bound_gradients}, no rescaling is bad for appropriately handling common phrases.
No rescaling is also bad for appropriately handling rare ones, as we can see by its connection to the infinity-norm and the top row of the table.
No rescaling of features makes it very difficult to prune perfect predictors.

\paragraph{Singleton predictors.}
As a special case, ``singleton predictors'' are those that appear only once in the entire corpus, and appear for a positive example.  
Normally if such a rare phrase appears once in a positive example, it can be pruned as described above.  
Regardless of $q$, in order to remove singletons that predict for a document with no other predictors, we must have $\regC \geq 2(1-\mu)$.

This can be generalized somewhat. 
Consider, for $q=2$, if the count of a phrase for a single positive document is $s$ and the count for $t$ negative documents for that same phrase are 1 each.  The $L^2$ normalizing constant is then
\[ Z = \sum_{i=1}^n c_{ki} = s^2 + t \]
and $x_{ki} = s/\sqrt{s^2+t} \approx 1$ for the single positive document and $x_{ki} = 1/\sqrt{s^2+t} \approx 1/s$ for the few negative documents.  
This is approximately the same as the singleton phrase circumstance, and will therefore be pruned as above.

Proper selection of the tuning parameter is a better approach for pruning than cutting by dropping phrases with low counts (ignoring computational issues), as it can also prune near-singleton phrases with high counts.  This circumstance indeed arises.  In a study of the Fat/Cats corpus, below, the word ``lion,'' used 8 times for a report involving a plague-ridden mountain lion corpse (positive example) and 9 times in various negative examples, was kept as a predictor in the final summary of ``Disease'' when $\regC$ was too low ($\regC=0.5$).

\subsection{Regularization with Cross-Validation}
The traditional form of selecting a tuning parameter is via cross-validation where some metric of predictive performance is optimized.
For example, in our context we could calculate the predicted labeling of set-aside documents and select $\regC$ such that the average squared distance between predictions and actual is minimized.
As we will see, this tends to give longer lists which can be less interpretable as more important signal can be buried amongst less-relevant terms.
Generally, predictive performance is not directly a measure of our primary focus: the interpretability and significance of the selected phrases.

\subsection{Regularization with early stopping and the elastic net}

The original Ifrim, et al method includes both a penalty term in the loss function but also regularizes by stopping before full convergence.  Different choices of $\regC$ do affect the resulting model, but early stopping is an easy way to obtain a list of specified length quite quickly (although if the list is non-sensical then this is obviously not a good move).  Ifrim's initial paper in fact has no penalty term in the loss function at all---their entire regularization is due to early stopping.    

The relationship of these forms of regularization is unclear; early stopping clearly has great computational advantages and there is no need for convergence checks; we simply stop when we have found enough features of interest.  
However, it is unclear, for example, whether this approach alone will successfully prune out rare phrases.

\section{Computational Comparisons}
We compare our CCS tool to three other methods in two studies.
The first compares running time and general characteristics of the final word lists, generally using default values and recommended approaches for setting tuning parameters.
The second compares prediction accuracy for the four methods.
In a third study, we also examine the CCS tool under different choices of $q$.

For our data, we use our Fat/Cats corpus and our ECAB corpus, both of which we describe more fully in the case studies section below.

\subsection{The Four Methods}
The main comparison method is multinomial inverse regression (MNIR) \cite{Taddy:2013tp}, a text regression method that is primarily designed to be distributed across many cores in order to be able to handle massive data.
It parallelizes the regression by conducting individual inverse regressions of the counts of phrases onto the feature space (which is in our case the binary labeling).  
It is regularized, giving a sparse feature vector.
The recommendation of MNIR, regarding tuning, is to use AIC or BIC penalization.  
We selected BIC because we are more interested in interpreting the covariates than in the quality of prediction, and BIC is known to be superior for model recovery as compared to AIC (see, e.g., \cite{speed1993model} or, more generally, \cite{Hastie:2003wp}).
We used the \verb|textir| \cite{Taddy:2013tp} package.
MNIR resulted in very long lists (of often more than a thousand words) so we truncated the list by taking the words with the top 1000 weights.  For practical use we would advocate greater regularization via, e.g., cross validation, to restrict the list further.

We also compare to a classic linear Lasso approach (see, e.g., \cite{Tibshirani:2004tz}).  Earlier work \cite{Jia:2014ko} shows that the gains from logistic over linear are minimal, and the computational expense is large.
We use the \verb|glmnet| package \cite{friedman2010regularization}, selecting the regularization parameter with cross validation on the RMSE test error.
The standard package automatically standardizes the columns, so this method uses $L^2$-rescaling.
Generally the Lasso lists were short, but if they were above 1000 words we truncated as above.

We finally compare to the original $n$-gram method of Ifrim et al.
Other than our own, there does not seem to be an R package implementing this approach, so we replicated their method by using our binary features option and not rescaling the columns.
For a more direct comparison, we select the tuning parameter with our permutation approach.
They initially advocated early stopping to regularize, but in later work (and in their C++ package which we extended and modified for our package) they introduced direct regularization.

In the first two studies we used $L^2$ normalization for our method.
We select $C$ two ways: the permutation approach discussed above and to a fixed $C=8$ to prune perfect predictors for 3 or fewer documents.
We did not allow gaps in phrases, and did not upper bound phrase length.
For the permutation, we permuted 10 times and took the maximum as the $C$ as the $C$ from the permutations do not vary much.

We ran our trials primarily on the cleaned Fat/Cat dataset stemmed with Porter stemming \cite{porter1980algorithm} via the \verb|tm| \cite{meyer2008text} package.
See Section~\ref{sec:data_cleaning} for further details.
For the latter two methods the data was stored in a flat file of cleaned and stemmed text, with each line corresponding to a single document.
For the first two methods we generated a document-term matrix from this text, dropping all terms that appeared fewer than 5 times to keep the matrix manageable.  
This resulted in 8698 unigrams, 81,863 bigrams and 122,528 trigrams.  
There are 49,558 documents in total.
We ran the lasso twice, first with unigrams and second with all unigrams through trigrams.
For MNIR we only use unigrams; the number of features generated when we expanded to trigrams was computationally prohibitive on a single computer.

To obtain labeling, we selected a random sample of 100 of the 1400 keywords associated with the Fat/Cat reports, weighted by the frequency of the keywords's appearance, to form the labeling schemes to evaluate.
For each keyword we dropped any phrases from consideration as a feature (e.g., phrases with ``carbon'' or ``monoxide'' for the carbon monoxide keyword were dropped) by either passing these words as banned words to our algorithm or dropping the relevant columns from the document-term matrix.

To further understand computational timing, we also replicated our first comparison study on the ECAB legal decisions discussed below.  
Here we had 7 judges that were part of overlapping subsets of decisions, and we compared each judge to baseline (this is not our labeling of primary interest in the case study below).  

For the first computational comparison, we ran the four methods on the full data and compared runtimes and characteristics of the final word lists.
For the second computation comparison, we set aside a random 15\% of the data, and then predicted the labeling of the set-aside data using each method. 
The Lasso, the Ifrim et al., and our method all produce nominal predictive scores, often overly low due to the massive imbalance of the positive and negative classes.
To find a cut-point for classification, we fit a logistic regression model on whatever predictive value came out of the method on the training set, and then classified by determining whether the predicted log-odds were above or below 0.
For the MNIR method we first projected the results of the inverse regression as described in \cite{Taddy:2013tp} and used the resulting score as the covariate in the logistic regression modeling step.
We also calculated AUC scores using the raw predictive scores for all methods.

Our third investigation was on the impact of different choices of $q$ for $L^q$-rescaling when using our method.
We again generated word lists for each of our 100 keywords, and calculated average length and average frequency of words for the resulting lists for $q = 1.2, 2,$ and $4$.
We put no upper bound on phrase length.

Simulations were run on a MacBook pro with a 2.9 GHz Intel Core i5 processor, 16 GB of memory, and a solid state drive.
Reproducible R code for all simulations is available on request.

\subsection{Comparison Results}

For each labeling and each method, we calculated different statistics on the word lists.
We also timed the total time to generate a list.
Table~\ref{tab:comparison_results} has the average of these measures across the 100 labelings.
A few general trends are evident.

\paragraph{List characteristics.} 
Overall the resulting lists are very different in character.
We calculated average list length, and then, for lists truncated at the top 1000 terms, calculated the mean and median frequency of the words to gauge how common or rare selected phrases were.
We finally averaged these means and medians across the different labeling runs.
We see the different methods are quite different in terms of these scores.

MNIR gives very long lists of very specific words; this is probably due to the inverse regression, which selects all phrases that are relevant without much regard to correlation structure between them.
The Lasso also tends to have longer lists, but the average appearance of the selected words is on par with the $n$-gram feature-rescaled regressions.
However, when the Lasso had access to trigrams, which can be highly targeted and specific, the median frequency plummeted to 13.  For pentagrams, it went to 12.
As expected, the Ifrim method, due to no column rescaling, selects \emph{very} general terms, with a median frequency of around 10,000.

For the OSHA dataset, the permutation $C$ values tended to be close to the fixed $C$, giving overall word lists that were similar as well.  
For the ECAB dataset, with longer documents, the permutation-selected $C$ was much higher, and thus the word lists were much shorter due to the greater regularization.  
The resulting words were also typically more general.

Overall, it is clear that the practitioner can use different methods to get different types of lists.
We believe, generally, that one wants short lists of phrases, and that those phrases should not be overly general.

\paragraph{Runtime.}
Runtimes were generally comparable for our OSHA data but widely different for the ECAB data.
We discuss the OSHA data first.
The Lasso method, even including its cross-validation step, was quite comparable to the textreg method with respect to time due to its very fast implementation.
It is also robust to very wide feature matrices; note the average time for using all trigrams is the same as for just unigrams.
The Ifrim method is faster than the textreg methods, likely due to improved pruning.
MNIR is also quite fast, event though it is designed for parallel systems and we ran on a single core.
Generally, the computational times to generate summaries are comparable.
However, MNIR on the trigrams was not workable.  As MNIR is linear in the number of features, the blow-up of features was too much of a time increase.
Of course, multiple machines and its parallel structure could avoid this.

We were quite surprised by the time statistics being insensitive to number of terms for the \verb|glmnet| package.  
To investigate further we calculated a document-term matrix for all phrases up to 5 words, giving 363,132 unique terms, giving the extra row in Table~\ref{tab:comparison_results}.  
Here average runtimes for the Lasso increased modestly by about 16\% to a mean of 101 seconds.

The computational times were more spread out for the ECAB data; see the bottom half of Table~\ref{tab:comparison_results}.
We now see a substantially increased time from moving to unigrams to trigrams for the Lasso and the timing of the textreg methods exploded.  
In investigating this time differential further, we found many selected phrases had 5 or more words.
Furthermore, most runs reached the maximum number of iterations before convergence, indicating flat surfaces. 
This is a weakness of greedy coordinate descent, which we discuss further below.
Overall we potentially have, due to the use of boilerplate language in legal writing, long yet informative phrases that we wish to see in our summaries.
This could make pre-generation of candidate phrases prohibitively expensive.
The MNIR method selected all unigrams; each unigram apparently has enough difference in use across judges to be selected under the BIC penalization.

The times for the Lasso and MNIR do not include the time to generate the document-term matrix, which was 7.4 minutes for 5-grams and 3.7 minutes for trigrams.

\begin{table}[ht]
\centering
\begin{tabular}{lrrrrr}

 &  time   & list length  & \multicolumn{2}{c}{word freq}  \\ 
method &  (avg sec)  & (avg) &  (avg) &  (median) \\ 
  \hline
  Ifrim & 21.5 & 25 & 15,285 & 10,178  \\ 
  lasso & 87.6 & 194   & 5,441 & 217  \\ 
  lasso (trigrams) & 86.6 & 440  & 657 & 13  \\
  lasso (pentagrams) & 101.1 & 502  & 503 & 12  \\ 
  MNIR & 42.1 & 8,696 & 3,818 & 392   \\ 
  textreg & 71.2 & 25 &  1441 & 371  \\ 
  textreg (fixed $C$) & 46.4 & 24  & 1711 & 472  \\ 
   \hline
     \hline
Ifrim & 63.8 & 8 & 17,861 & 16,872 \\ 
  lasso & 38.5 & 261 & 3643 & 109 \\ 
  lasso (trigrams) & 138.8 & 800 & 898 & 21 \\ 
  MNIR & 163.9 & 25,990 & 31 & 16 \\ 
  textreg & 540.3 & 14 & 12,595 & 10,639 \\ 
  textreg (fixed $C$) & 555.6 & 64 & 2481 & 653 \\ 
   \hline
\end{tabular}
\caption{Results of comparison study on OSHA reports (top) and ECAB legal decisions (bottom).  Runtime does not include time to generate the document-term matrix for the Lasso and MNIR.} 
\label{tab:comparison_results}
\end{table}

\subsection{Predictive Accuracy Results}

\begin{figure}[htbp]
\begin{center}
\includegraphics[width=0.8\textwidth]{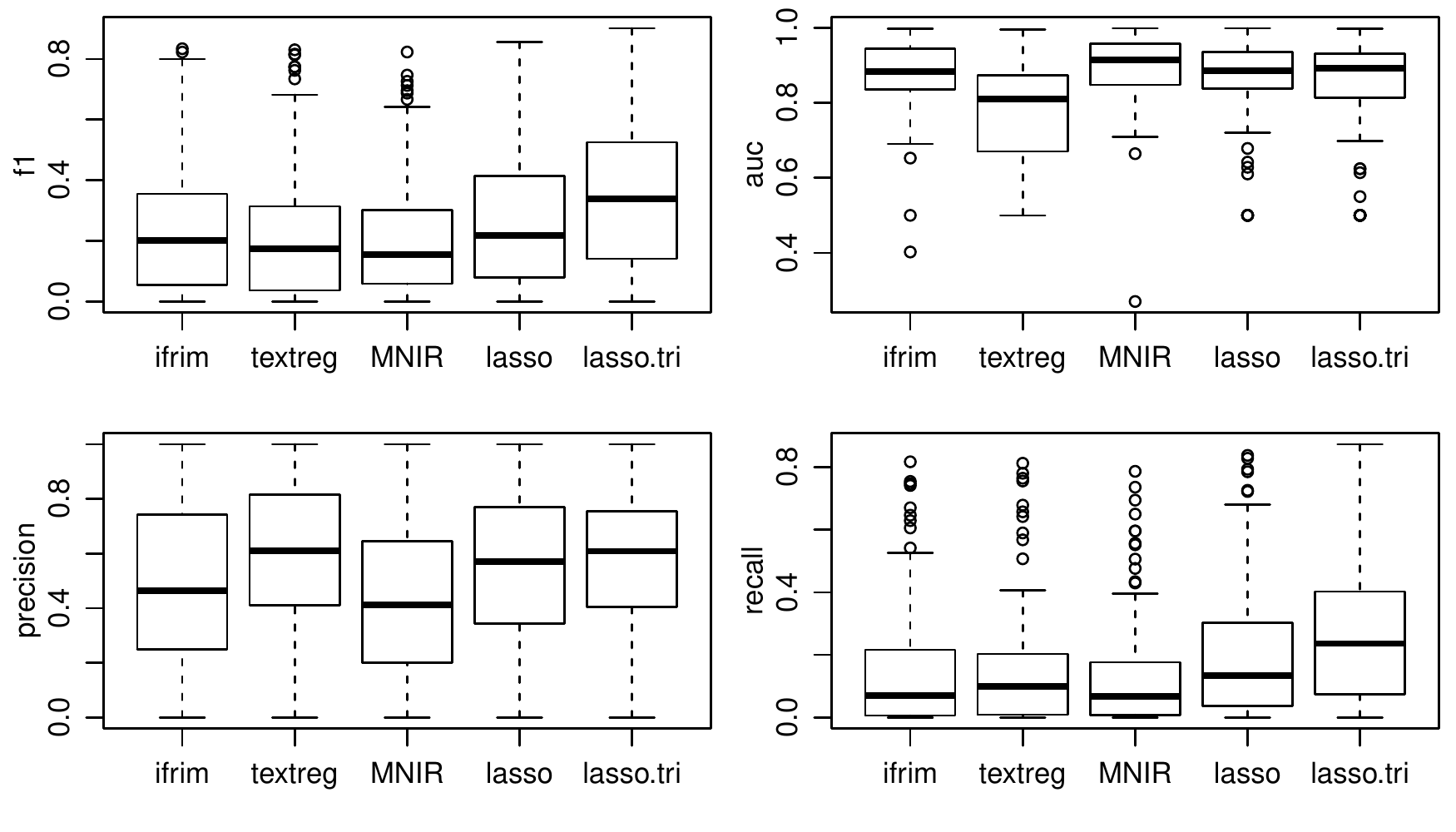}
\caption[Out-of-sample classification rates for different methods]{Out-of-sample classification rates for different methods.  Lasso on trigrams (Lasso-3) is generally the best, although there is substantial variability.  Keywords where any of the methods failed to return $F1$ scores were dropped.  Charts substantively the same when these keywords included.}
\label{fig:recall_scores}
\end{center}
\end{figure}

To assess predictive accuracy we used the macro-averaged $F1$ statistic \cite{yang1999re} ($F1$ calculated for each of the 100 trials and then averaged).  
See Table~\ref{tab:recall_results}.
All methods would sometimes score none of the test set documents as positive, giving undefined $F1$ scores.  These are also indicated on the table.
Figure~\ref{fig:recall_scores} shows these scores along with the component precision and recall, as well as AUC scores, for those keywords where all methods had defined $F1$.
The predictive accuracy is comparable across the methods, although our text regularization does suffer some due to over-regularization from the permutation method.
Overall, predictive accuracy is low; recall, in particular, tends to run around 20\% for all methods, although the Lasso with trigrams was noticeably superior.

We also examined AUC scores.  
Here we again see the cost of textreg's over-regularization: conservative classification reduces sensitivity greatly, lowering the ROC and AUC scores.
The AUC scores are deceptively high due to the imbalance between positive and negative examples.

Generally, the superior performance of the Lasso on trigrams indicates that rescaling features coupled with the richer feature set of multiple words is useful for prediction tasks.
Our methods results indicate over-regularization is detrimental to prediction due to, primarily, diminished ability to predict positive documents as positive (shown by the recall scores).

\begin{table}[ht]
\centering
\begin{tabular}{lrrr}
  \hline
method & f1 & sd & \# missing \\ 
  \hline
MNIR & 0.22 & 0.22 & 19 \\ 
  lasso & 0.28 & 0.24 & 6 \\ 
  lasso (trigrams) & 0.36 & 0.25 & 6 \\ 
  Ifrim & 0.25 & 0.24 & 16 \\ 
  textreg & 0.23 & 0.23 & 7 \\ 
   \hline
\end{tabular}
\caption{$F1$ scores for different methods averaged across keywords examined (with standard deviations).  Number of keywords with no defined $F1$ (out of 100) also indicated.}
\label{tab:recall_results} 
\end{table}

\subsection{Selection of $q$ Results}

As anticipated, different values of $q$ produce lists of different quality.  Results are on Table~\ref{tab:comparison_results_q}.
Higher $q$ corresponds to lists with relatively more common words and phrases.
Low $q$ produces lists that tend to have phrases with more words.  For the $q=1.2$ list, more than half the phrases were 3 words or longer.
We also found that the length of the list increased with $q$.  For $q=4$ we had longer lists with more common words.  For low $q$ the algorithm selected very targeted phrases, and not too many of them.
The regularization parameter $C$ is not comparable across $q$.  
Therefore we recalculated it via the permutation approach for each run and value of $q$.
Mean values for $C$ are shown on the table for reference.

Generally we find a rule of thumb worth remembering: longer lists tend to have more general terms.  
This can happen by adjusting either $C$ or $q$ (see first simulation and compare different $C$, for example).

\begin{table}[ht]
\centering
\begin{tabular}{lrrrrrr}
  \hline
method &  time & list length & phrase length & word freq  & word freq  & C  \\ 
 &  (avg sec) &  (avg) &  (avg) &  (avg) &  (median) &  (avg) \\ 
  \hline
textreg.1.2 & 78.7 & 8 & 2.9 & 310 & 110 & 4.6 \\ 
  textreg.2 & 72.8 & 24 & 2.2 & 1492 & 435 & 7.1 \\ 
  textreg.4 & 135.3 & 35 & 1.8 & 8035 & 1761 & 19.0 \\ 
   \hline
\end{tabular}
\caption{Word list characteristics of different rescaling norms.  Length is length of list, mean size is average number of words in the selected phrases, frequency refers to phrase occurrence in the corpus.  Mean $C$ is the average regularization value.} 
\label{tab:comparison_results_q}
\end{table}

\subsection{Discussion}

Overall, our method succeeds in accommodating multiple word phrases while also allowing for an intercept and the rescaling of features.
As shown, these extensions are critical for generating manageable lists with phrases that are not overly common.
Further, picking $C$ and $q$ does control both list length and commonality of words on the lists as predicted by our initial discussion.

In terms of computational time our method did not perform particularly well when compared to the Lasso of the \verb|glmnet| package, especially considering that the Lasso's overall time includes that of cross validation. 
As our method is similar to Lasso regression on a pre-constructed phrase matrix (differing only in that it uses a hinge-loss instead of a squared loss), one might naturally ask whether an alternative approach would be to simply generate the full document-term matrix and use \verb|glmnet|.
We are not entirely convinced, as we discuss next.

First, generating full document-term matrices including phrases with wildcards could be computationally tedious, especially if we imagine extensions such as optionally included words.
At the very least, it is expensive in both time and memory, and grows increasingly so with the number of possible phrases; the 7 minutes to generate all 5-grams is well over the model-fitting times.  
The stored matrix was twice the size of the initial raw text.
Second, the hinge loss is different from the classic $L^2$ loss, and there could be core differences in overall behavior here; this is an area for future investigation.
Third, the speed of the \verb|glmnet| package is partially due to the LARS approach where the entire regularization path is computed at once.
It is arguably also due to the package being particularly well-developed and optimized for efficiency.
There may be similar ways to substantially speed up our package and method to make it more competitive.

For example, one potential slow-down, determined from examining the convergence paths of many runs, is that we often see an initial selection of a small list of phrases and then a slow hill climb where with each step a different already-selected phrase is selected for adjustment.
This comes from the coordinate descent; the optimal gradient path is at an angle, and so following it requires small steps in the associated coordinate directions to trace that path.
Unfortunately, with each such step the algorithm conducts a full search for the highest gradient across all phrases, which is expensive.
We might instead, after each phrase is selected, find the maximum point given the set of all phrases selected at that point.  
Then only if a new phrase were introduced would the algorithm have further steps.
This is effectively the LARS \cite{Tibshirani:2004tz} approach, and is an important area for enhancement.

Furthermore, the greedy ascent iteration often involves the intercept ratcheting down as features ratchet up.
This suggests another possible direction of forcing the intercept to be $-1$ (for corpora with rare positive labeling rates) rather than estimating it; a $-1$ intercept corresponds to predicting all documents as negative by default, distinct from no intercept which predicts 50-50 uncertainty as default or allowing the intercept to move which tends to predict the overall base rate as default.
Specifying the intercept value would save an intercept update with each step; the optimization problem is then finding a collection of phrases to give positive documents positive weight without affecting the negative documents too heavily.

Regardless, for truly massive data, especially when the number of documents grows large, neither our approach or the Lasso will work.
Instead, methods that allow for easy parallelization, such as MNIR, will be key.
Another area for future exploration, therefore, is to determine how to over-regularize MNIR to get shorter lists, which also might induce lists with more common phrases.

\section{Case Studies}

We illustrate the CCS tool with two case studies.
For the first case study, we also compare the resulting summaries from CCS to the three other methods discussed above.
An overview of the code to generate these results using our textreg R package (available on CRAN) is in the appendix.
Full scripts and data are available on our website.\footnote{Website address is \url{http://scholar.harvard.edu/lmiratrix/software/textreg-r-package}}
For other studies using similar tools see, for example, \cite{Jia:2014ko} or \cite{ElGhaoui:2011uo}.
Before presenting our results, we discuss data representation.

\subsection{Data representation and cleaning}
\label{sec:data_cleaning}

There are many choices one might make in preparing a corpus for statistical analysis.
It is common to, for example, convert text to lowercase and to drop all punctuation. 
We take that approach here, although we convert all digits to `X' (to preserve the presence of numbers, in case that is informative) and convert hyphens to  blank spaces (so the sequence of hyphenated words would coincide with a non-hyphenated similar phrase---something not possible with single word analyses).
 
Most text analysis packages would then convert the raw text into an $m \times p$ matrix of counts, dropping any stop words, but because of the greedy coordinate descent algorithm, unknown phrase length, and the related generation of features on-the-fly, we store the text as raw strings, with one string per document.

There is some controversy as to whether to ``stem'' documents,  which is where the tails of many words are cropped so as to collapse the number of possible words. 
For example ``clean,'' ``cleaning'' and ``cleaner'' might all be cropped to ``clean+''
This has the disadvantage of making resulting text output somewhat difficult to read, especially when considering phrases.
Stemming can also lose textual meaning if the different suffixes are in fact important in the context.
It has the advantage, particularly for phrases, of collapsing several different versions of phrases into one.
We provide stemming as an option but, to enhance readability of output, append a `+' to the end of all stemmed words (and their roots) to indicate they have been potentially cropped.
We also provide tools to wildcard search for stemmed phrases in the original text so as to recover examples of the complete phrases.
 
Sometimes a given context involves words that are known a priori, or nearly a priori, to have no meaning.
We therefore provide a option for custom-made, short stop-word lists (i.e., a list of banned words) that are prohibited from being in any summary phrase.
Generally these lists are built on the fly as an iterative process. 
The first summary generated will often contain words that are immediately recognizable as inconsequential to a researcher with pre-existing contextual knowledge, even though they are correlated with the labeling.
The researcher would then drop these words, rerun the algorithm, and repeat as necessary.
We do not see any way to avoid this; the case studies below illustrate why.

\subsection{Fat/Cats}

Our main investigation relies on OSHA's publicly available summaries of occupational fatalities and catastrophes (``Fat/Cats''),\footnote{Website address is \url{http://ogesdw.dol.gov/views/data_catalogs.php}.} in the United States, from 2000 to 2010. These summaries describe workplace incidents that have resulted in death or the inpatient hospitalization of three or more workers. When such an event occurs, an employer must report it to OSHA.\footnote{29 C.F.R. 1904.39(a)}\footnote{OSHA recently expanded the list of reportable events to include the loss of an eye, amputation, or inpatient hospitalization. Occupational Injury and Illness Reporting Requirements-NAICS Update and Reporting Revisions, 79 Fed. Reg. 56129-56188 (September 18, 2014)(amending 29 C.F.R. 1904.39).}
In the course of conducting the resulting investigation, OSHA generates a narrative report, part of which becomes publicly available and is annotated with any of a set of about 1400 keywords to categorize the narrative reports in terms of specific chemicals involved, machinery involved, body parts affected, and other salient features.
The publicly available records primarily consist of a title, a short paragraph summary of the incident, along with the date, whether the incident involved a fatality, and several other covariates. 

We concatenated the title and paragraph description to form the documents.  These documents tend to be
56 to 136 words long (these are the 1st and 3rd quantiles), with a minimum length of 4 words and a longest report of 791 words.  After stemming there were 49,840 unique unigrams (word stems), of which 12,700 appeared 10 or more times and 4,704 appeared 100 or more times.

To investigate this corpus, we can, for any keyword, generate a labeling of the narrative reports by setting those reports tagged with the keyword as $+1$ and the remainder as $-1$.  
Using CCS on this would then summarize the collection of reports marked with a keyword by comparing them to all other reports.
Ideally this would take out words common to these reports (e.g., ``employee'' or other general work-place terms), leaving us with phrases that make the identified set stand out.
We would interpret this summary as a distillation of what is distinct about this category of Fat/Cats as compared to Fat/Cats in general.  
By periodically summarizing reports for each keyword of interest, researchers may gain information about emerging hazards and trends in circumstances. 
Hopefully the resulting summaries would be faster to read than the individual narratives, but still contain hints as to general themes within these narratives.

As chemical exposure is an area of particular interest for enhanced surveillance and understanding, we generated a background comparison set of documents by identifying keywords that we deemed to be at least loosely associated with chemical exposure. We then defined the ``chemical family'' of narratives as all narratives that were labeled with at least one of these keywords. This allowed us to compare various categories of narratives within the limited context of this chemical family, as well as within the larger context of all other types of narratives. Changing the background set highlights different aspects of what sets apart a marked collection of reports.

As an overview of the overall number of narratives of different topics of interest, Table~\ref{tab:label_counts} shows the appearance pattern of the categories examined.  We discuss Methylene Chloride and Carbon Monoxide here, and defer Chemical Reaction to a supplementary document.
The table also shows how many narratives involved a fatality.

\begin{table}[ht]
\centering
\begin{tabular}{rrr|rr}
  & \multicolumn{2}{c|}{General} & \multicolumn{2}{c}{Chemical} \\

             & \# & \% & \# & \%  \\ 
  \hline
Methylene Chloride &  17 & 0.03\% &  17 & 0.2\% \\ 
  Carbon Monoxide & 243 & 0.5 & 243 & 3.5 \\ 
  Chemical Reaction & 115 & 0.2 & 115 & 1.6 \\ 
  \hline
  Fatality & 20,691 & 41.8 & 2,575 & 36.7 \\ 
  Total    & 49,558 & 100.0 & 7,014 & 100.0 \\ 
  \hline
\end{tabular}
\caption[Number of Narratives with Different Keyword Labelings]{Number of Narratives with Different Keyword Labelings.  Second pair of columns restrict database to only reports related to chemical exposure.} 
\label{tab:label_counts}
\end{table}

\subsubsection{Methylene Chloride}

As it is our motivating example, we first examined Methylene Chloride.  
We initially selected a value of $\regC=4$ to ensure that we prune all singleton perfect predictors (see Section~\ref{subsec:prune}).
There are 17 reports marked with the ``Methylene Chloride'' keyword.  
Running CCS on these reports returns two words, ``methylene'' and ``chloride.''
As these words are not of interest, we immediately added these words to the ban list and reran.
Table~\ref{tab:meth_chl} displays the resulting summary comparing these 17 narratives to all the other narratives. 

The summarizer picks up on the coherent theme across these reports of bathroom refinishing.
This example is quite encouraging, given our prior knowledge of the dangers of Methylene Chloride, but the utility of CCS in detecting yet unknown patterns remains to be seen.
If we select $\regC$ based on the 95th percentile of 100 permutations, we obtain $\regC=5.65$.  
The needed $\regC$ to result in a null summary is, by comparison, $\regC^{obs} = 6.92$.
We conclude that there is a statistically significant relationship between the text and the keyword (beyond the presence of the banned words), and that the summary is thus informative.
The corresponding summary for $\regC=5.65$ is quite succinct, containing only ``a bathtub'' and ``stripper contained.''
Picking $\regC$ to give statistical significance appears, here, to drop informative phrases.  

\begin{table}[ht]
\centering
\begin{tabular}{lrrrrr}
  \hline
phrase & \# phrase & \# reports & \# tag &  \% tag &  \% phrase \\ 
  \hline
  a bathtub & 9 & 9 & 5 & 56 & 29 \\ 
  paint stripper & 5 & 5 & 3 & 60 & 18 \\ 
  stripper contained & 3 & 3 & 3 & 100 & 18 \\ 
  and reglazing & 2 & 2 & 2 & 100 & 12 \\ 
  from a bathtub & 2 & 2 & 2 & 100 & 12 \\ 
  remover contained & 2 & 2 & 2 & 100 & 12 \\ 
  stripping agent & 2 & 2 & 2 & 100 & 12 \\ 
  tub head & 2 & 2 & 2 & 100 & 12 \\ 
   \hline
\end{tabular}
\caption[Methylene Chloride Results]{Methylene Chloride ($L^2$-rescaling, against all reports).  \# Phrase is total occurrence of phrase; \# reports is number of reports containing phrase; \#  tag is number of Methylene Chloride reports containing phrase; \% tag is percent of phrase appearances in Methylene Chloride reports; and \% phrase is percent of Methylene Chloride reports containing phrase. } 
\label{tab:meth_chl}
\end{table}

The summaries do not necessarily capture information in all tagged documents.
In this case, for example, six of the Methylene Chloride reports do not have any of the phrases on Table~\ref{tab:meth_chl}, and so are not represented.
A manual review of these reports revealed that four involved ``stripper''s for tile, floors, and furniture.  
One involved an explosion and one, quite terse, only referred to Methylene Chloride gas.

\subsubsection{Carbon Monoxide}
We also examined reports relating to Carbon Monoxide, an asphyxiant odorless gas.  
We ran CCS with different values of $q$ for $L^q$-rescaling to examine the impact of different levels of rescaling.  
We compared the CO cases to all other cases involving any of a set of keywords predetermined to be related to some sort of chemical exposure (i.e., those narratives marked as members of the ``chemical family'').
To reduce computational time, we limited attention to phrases that appear at least five times in the corpus.
Results are in Table~\ref{tab:CO}.  

To obtain these results we summarized in an iterative process; words such as ``carbon,'' ``monoxide,'' ``gas,''  ``poisoning,''  ``exposed,''  ``exposure,''  ``overexposed,'' ``carboxyhemoglobin,'' ``ppm,'' ``levels,'' 	``partspermillion,'' ``overcome,'' and ``co'' were eventually dropped.  
We also removed the more specialized ``hyperbaric,'' having to do with a medical intervention for CO poisoning and ``cohb,'' an abbreviation for carboxyhemoglobin, a molecular complex that hemoglobin and carbon monoxide form in the body.
All of these words appeared in initial summaries and are due to the technical and/or obvious aspects of CO poisoning; they do not reveal trends or characteristics of interest and thus obscure the desired results.
None of these words would have appeared on any conventional stop word list.
As they are in fact correlated with the category, we see no way of automatically removing them.

The final results reflect several known patterns in CO poisoning.  
For example, the exhaust from gasoline and propane powered engines are major culprits of these exposures, particularly in combination with poorly ventilated enclosed spaces. The appearance of the phrase ``cold room'' appears to reflect incidents in which propane-fueled forklifts and floor cleaning devices were the source of carbon monoxide exposure within cold storage areas, where ventilation can be poor.  

In investigating hospitalization, we found 17\% of the CO poisoning cases contained ``were hospitalized'' versus only 5\% of the other chemical-related narratives and 1\% of the non-chemical narratives.
The fire department was mentioned in nearly 9\% of these narratives versus a baseline of 0.5\%.
This all may be due to lower rates of fatality, with only 36\% of the CO poisoning cases involving fatalities as compared to 37\% for other chemical family reports and 43\% for non-chemical-family reports.
Interestingly, ``dead'' appears in 16\% of the CO narratives as compared to 5\% in the remainder of the chemical family.

Different rescalings give different styles of summaries. 
The smaller $q=1.2$ and $q=1.5$ have very specific phrases (e.g., ``were using a gasoline'' and ``their blood'') that appear only in the positively marked documents.
Larger $q$ give more phrases overall, and give phrases that appear at higher rates in both the positive and negative class.  
For example, ``employees,'' with more than 10,000 appearances, appears for $q=4$.

Overall, infrequent and specific phrases are relatively easy to interpret, and the more common phrases less so.  
But their patterns of appearance are striking.  
``Employees,'' for example, appears in 52\% of the CO narratives versus a baseline of a mere 11\%!
Less surprisingly, ``enclosed'' appears 7\% of the time versus less than 0.5\% at baseline.

Table~\ref{tab:CO} contrasts ``Carbon Monoxide'' to all incidents labeled with other chemical-related keywords.
We also compared CO cases to the full set of cases in the database.
That is, we summarized the same collection of reports, but used a different baseline point of comparison.
Results are in Table~\ref{tab:CO_full}, in the Appendix.  
They are broadly similar.

We also analyzed the data using stemming.
See Table~\ref{tab:CO_stem} in the Appendix.  
Results are again broadly similar (but possibly harder to read).  
Stemming collapses phrases, which can be helpful, but hampers human readability.

\begin{table}[ht]
\centering
\tiny
\begin{tabular}{lrrrrrrrrr}
  \hline
phrase & q=1.2 & q=1.5 & q=2 & q=3 & q=4 & $\#$ reports & $\#$ tagged & \% tagged & \% phrase \\ 
  \hline
a propane powered & 0.25 & 0.02 &  & 0.05 & 0.07 &  17 &  15 & 88.00 & 6.00 \\ 
  their blood & 0.16 & 0.13 &  &  &  &   5 &   5 & 100.00 & 2.00 \\ 
  gasoline powered & 0.15 & 0.29 & 0.28 & 0.25 & 0.21 &  41 &  30 & 73.00 & 12.00 \\ 
  concrete saw & 0.11 & 0.24 & 0.18 & 0.17 & 0.08 &  10 &   9 & 90.00 & 4.00 \\ 
  an X hour & 0.10 &  &  &  &  &   6 &   6 & 100.00 & 2.00 \\ 
  the fire department measured & 0.09 &  &  &  &  &   4 &   4 & 100.00 & 2.00 \\ 
  in a cold & 0.09 &  &  &  &  &   4 &   4 & 100.00 & 2.00 \\ 
  operating a propane & 0.09 & 0.13 &  &  &  &   6 &   6 & 100.00 & 2.00 \\ 
  propane operated & 0.07 &  &  &  &  &   4 &   4 & 100.00 & 2.00 \\ 
  were using a gasoline & 0.07 &  &  &  &  &   5 &   5 & 100.00 & 2.00 \\ 
  propane powered &  & 0.25 & 0.23 & 0.22 & 0.13 &  34 &  28 & 82.00 & 12.00 \\ 
  powered &  & 0.20 & 0.35 & 0.41 & 0.44 & 169 &  85 & 50.00 & 35.00 \\ 
  forklifts &  & 0.13 & 0.21 & 0.26 & 0.22 &  23 &  16 & 70.00 & 7.00 \\ 
  for fresh &  & 0.13 &  &  &  &   4 &   4 & 100.00 & 2.00 \\ 
  the generator was &  & 0.06 &  &  &  &   9 &   7 & 78.00 & 3.00 \\ 
  X hour &  & 0.02 & 0.03 &  &  &  16 &  11 & 69.00 & 5.00 \\ 
  overexposure &  &  & 0.15 & 0.26 & 0.28 &  37 &  18 & 49.00 & 7.00 \\ 
  exhaust &  &  & 0.04 & 0.10 & 0.10 & 112 &  35 & 31.00 & 14.00 \\ 
  generator was &  &  & 0.04 & 0.08 &  &  14 &  10 & 71.00 & 4.00 \\ 
  blood &  &  & 0.03 & 0.06 & 0.04 &  98 &  30 & 31.00 & 12.00 \\ 
  a gasoline &  &  &  & 0.08 & 0.10 &  48 &  26 & 54.00 & 11.00 \\ 
  average &  &  &  & 0.07 & 0.05 &  16 &  10 & 62.00 & 4.00 \\ 
  found &  &  &  & 0.06 & 0.06 & 554 &  68 & 12.00 & 28.00 \\ 
  were treated &  &  &  & 0.05 & 0.05 & 111 &  22 & 20.00 & 9.00 \\ 
  the fire department &  &  &  & 0.05 & 0.01 &  90 &  21 & 23.00 & 9.00 \\ 
  hour &  &  &  & 0.03 & 0.03 &  85 &  23 & 27.00 & 9.00 \\ 
  source of the &  &  &  & 0.01 & 0.02 &  32 &  12 & 38.00 & 5.00 \\ 
  ventilation &  &  &  & 0.01 & 0.02 & 130 &  32 & 25.00 & 13.00 \\ 
  enclosed &  &  &  & 0.01 & 0.01 &  66 &  18 & 27.00 & 7.00 \\ 
  were taken &  &  &  & 0.01 & 0.02 &  97 &  23 & 24.00 & 9.00 \\ 
  employees &  &  &  & 0.01 & 0.01 & 1400 & 126 & 9.00 & 52.00 \\ 
  employees were &  &  &  & 0.01 & 0.01 & 590 &  66 & 11.00 & 27.00 \\ 
  were &  &  &  & 0.00 & 0.01 & 2777 & 168 & 6.00 & 69.00 \\ 
  fire department &  &  &  &  & 0.02 & 211 &  36 & 17.00 & 15.00 \\ 
  a propane &  &  &  &  & 0.02 &  96 &  26 & 27.00 & 11.00 \\ 
  were hospitalized &  &  &  &  & 0.01 & 384 &  42 & 11.00 & 17.00 \\ 
  dead &  &  &  &  & 0.01 & 372 &  38 & 10.00 & 16.00 \\ 
  warehouse &  &  &  &  & 0.01 &  92 &  19 & 21.00 & 8.00 \\ 
   \hline
\end{tabular}
\caption[Different Summaries of Carbon Monoxide]{Different Summaries of Carbon Monoxide (compared to Chemical Family narratives)} 
\label{tab:CO}
\end{table}

Finally, we compare CCS to the other methods of MNIR, the Lasso, and the Ifrim et al. approach.
The methods returned summaries of very different lengths: MNIR was 293, Lasso 65, Ifrim 20, and textreg 12.
We had to truncate  MNIR and Lasso to display the lists, but we did this by taking the union of the top 20 words of each list and then displaying weights before truncation to see maximal overlap.
See Table~\ref{tab:CO_meth_compare}, with words sorted by frequency of appearance in the corpus.

We see the lists are quite different with mild overlap.
MNIR generally targets rare phrases, most of which are not displayed, Ifrim very general ones.  
MNIR, restricted to unigrams in this instance due to computational concerns, has less overlap than it might otherwise.
Textreg has mid-range phrases with a few very rare phrases which are perfect predictors.  
These phrases were not included in the document-term matrix due to their rarity, so could not be on the Lasso or MNIR lists.
Many of the phrases have similar meanings. 
For example, Ifrim has the general versions (e.g., ``gasolin+'') for the more specific (``gasolin+ power+'')  of the Lasso and textreg.

\begin{table}[ht]
\tiny
\centering
\begin{tabular}{lrrrrr}
  \hline
phrase & MNIR & lasso & ifrim & textreg & num.phrase \\ 
  \hline
power+ generat+ to+ &  &  &  & 0.4 & 3 \\ 
  level+ of approxim+ &  &  &  & 0.4 & 4 \\ 
  food+ poison+ &  & 0.1 &  &  & 5 \\ 
  for fresh+ &  & 0.3 &  &  & 5 \\ 
  propan+ power+ floor+ &  & 0.2 &  &  & 5 \\ 
  tailpip+ & 6.6 &  &  &  & 5 \\ 
  their blood &  & 0.2 &  &  & 5 \\ 
  unventil+ & 5.6 &  &  &  & 5 \\ 
  vanguard & 5.6 &  &  &  & 5 \\ 
  over+ expos+ to+ &  & 0.2 &  &  & 6 \\ 
  decatur & 6.2 &  &  &  & 7 \\ 
  receiv+ oxygen+ &  & 0.1 &  &  & 7 \\ 
  mek & 5.2 &  &  &  & 8 \\ 
  twa & 7.2 &  &  &  & 8 \\ 
  carbonyl & 7.3 &  &  &  & 9 \\ 
  exhaust+ fume+ &  & 0.1 &  &  & 9 \\ 
  intern+ combust+ engin+ &  & 0.2 &  &  & 9 \\ 
  newton & 5.5 &  &  &  & 9 \\ 
  power+ concret+ &  & 0.1 &  &  & 11 \\ 
  qa & 5.4 &  &  &  & 11 \\ 
  transient & 5.4 &  &  &  & 11 \\ 
  stratton & 5.6 &  &  &  & 12 \\ 
  three employe+ are expos+ &  &  &  & 0.3 & 12 \\ 
  weight+ averag+ &  & 0.1 &  &  & 12 \\ 
  brigg+ & 5.4 &  &  &  & 13 \\ 
  emiss+ & 5.4 &  &  &  & 13 \\ 
  fd & 6.2 &  &  &  & 14 \\ 
  poison+ at+ &  & 0.2 &  & 0.2 & 17 \\ 
  employe+ overcom+ by &  & 0.1 &  &  & 27 \\ 
  overexpos+ & 5.9 & 0.2 &  & 0.5 & 51 \\ 
  overexposur+ & 5.1 & 0.1 &  & 0.1 & 55 \\ 
  propan+ power+ &  & 0.2 &  & 0.3 & 56 \\ 
  gasolin+ power+ &  & 0.2 &  & 0.3 & 74 \\ 
  overcom+ by &  & 0.1 &  & 0.1 & 207 \\ 
  poison+ & 5.8 & 0.2 & 0.9 & 0.5 & 210 \\ 
  overcom+ & 4.3 &  & 0.1 &  & 249 \\ 
  ventil+ & 3.5 &  & 0.0 &  & 321 \\ 
  gasolin+ & 3.2 &  & 0.1 &  & 412 \\ 
  expos+ to+ &  & 0.0 &  & 0.0 & 490 \\ 
  exposur+ & 4.1 & 0.1 & 0.2 & 0.1 & 519 \\ 
  exhaust+ & 3.1 &  & 0.0 &  & 565 \\ 
  propan+ & 2.9 &  & 0.1 &  & 620 \\ 
  expos+ & 2.8 &  & 0.0 &  & 1053 \\ 
  level+ & 2.1 &  & 0.0 &  & 3282 \\ 
  found+ & 1.7 &  & 0.0 &  & 3502 \\ 
  power+ & 1.4 &  & 0.0 &  & 5779 \\ 
  forklift+ & 0.6 &  & 0.0 &  & 9118 \\ 
  fell+ &  &  & 0.0 &  & 19483 \\ 
  were+ & 1.2 &  & 0.0 &  & 27668 \\ 
  when &  &  & 0.0 &  & 36629 \\ 
  his &  &  & 0.0 &  & 53632 \\ 
  was &  &  & 0.0 &  & 176175 \\ 
   \hline
\end{tabular}
\caption{Different Summaries of Carbon Monoxide (comparing different methods)} 
\label{tab:CO_meth_compare}
\end{table}

\subsection{Legal decisions}

In the context of legal decisions, our motivating question is whether we can efficiently learn about characteristics of certain types of cases by extracting associated phrases and topics from a corpus of those cases. As an exploratory case study, we chose to examine publicly available decisions from the Employees' Compensation Appeals Board (ECAB), which considers appeals to determinations by the Office of Workers' Compensation Programs (OWCP) in the U.S. Department of Labor (DOL). OWCP handles compensation claims from federal workers injured during the course of employment. EBAC handles as many as 2000 appeals per year.

Within this case law, one particular area of interest is how ECAB handles compensation claims for so-called ``emotional conditions.''
These cases can be challenging for a number of interesting reasons. For example, establishing whether an employee's condition was legitimately caused by workplace conditions requires an analysis of causation that is unique in many ways from that which is appropriate in the context of a physical condition.

To further probe the potential utility of CSS in extracting useful information from large bodies of technical text, we performed an exploratory analysis on the collection of ECAB decisions that relate to both mental health conditions and causality. We sought to determine whether automated summaries would reveal meaningful patterns. 
These decisions are publicly available through DOL\footnote{Website address is \url{http://www.dol.gov/ecab/decisions/main.html}}.  
We examined years 2005 to 2010 by  scraping them from the website.

We ended up with 11,214 legal decisions, documents generally ranging in length from 1602 to 2691 words (these being the 1st and 3rd quartiles) and a median length of 2074 words.  The shortest was 281 words and the longest 12,664.  
There are 107,302 unique words, of which 37,474 appear 10 or more times and 11,264 appear 100 or more times.
These counts include case identifiers and other character strings as words.
We do not attempt to remove them directly.

 We  automatically labeled all of the decisions with two sets of dummy variables, one for emotional condition and one for discussion of causality or work-relatedness of the injury.
For each, we labeled documents if they contained any of a set of handpicked key phrases. Once we tuned our collection of key phrases we took a random sample of the positively and negatively marked documents and conducted a manual review.
The labeling is clearly not perfect, as is illustrated in Table~\ref{tab:ecab_sensitivity}.
Ideally, the CCS method will still be able to produce relevant summaries despite the noise of the missed labels.
Although it is possible that specific types of positive decisions are systematically missed due to the ad hoc labeling, discovery of meaningful summary phrases would nevertheless be suggestive.

\begin{table}[htp]
\begin{center}
\small
\begin{tabular}{lllllll}
			&			 &	\% Correct		&	\multicolumn{4}{c}{Estimated}		\\
Labeling   &Total	&   (sample) 	& Positive &Negative & Sense. & Spec. \\
			\hline 		
``Emotional Condition'' & 1479 & 94/100 & 1390 & 89 & 0.83 & 0.99 \\
No ``Emotional Condition'' & 9735 & 97/100 & 292 & 9443\\
Total Cases & 11214 \\
\hline 
``Causality/Work-Relatedness''  & 4236 & 99/100 & 4194 & 42 & 0.75 & 0.99 \\
No ``Causality/ Work-Relatedness'' & 6978 & 80/100 & 1396 & 5582 \\
Total Cases & 11214 \\
\hline
\end{tabular}
\end{center}
\caption{Manual Review of Labeling Quality for Legal Decisions}
\label{tab:ecab_sensitivity}
\end{table}

\begin{table}[ht]
\small
\centering
\begin{tabular}{rrrrrr}
  \hline
 & main & +``ecab'' & +``depression'' & +``xx'' & +many \\ 
  \hline
illness that is & 0.46 & 0.48 & 0.48 & 0.60 & 0.60 \\ 
  cutler xx ecab & 0.32 &  &  &  &  \\ 
  requirement imposed by the & 0.31 & 0.31 & 0.31 & 0.32 & 0.32 \\ 
  psychiatrist & 0.05 & 0.05 & 0.05 & 0.06 & 0.06 \\ 
  incidents alleged to have caused & 0.05 & 0.05 & 0.05 &  &  \\ 
  lillian cutler & 0.04 & 0.03 & 0.03 &  &  \\ 
  compensable & 0.03 & 0.03 & 0.03 & 0.03 & 0.03 \\ 
  disorder & 0.01 & 0.01 & 0.01 & 0.01 & 0.01 \\ 
  factor of employment and & 0.01 &  &  & 0.01 & 0.01 \\ 
  a factor of employment & 0.01 & 0.01 & 0.01 & 0.02 & 0.02 \\ 
  not covered & 0.01 & 0.01 & 0.01 & 0.03 & 0.03 \\ 
  reaction to & 0.01 & 0.00 & 0.00 &  &  \\ 
  anxiety & 0.00 & 0.00 & 0.01 & 0.01 & 0.01 \\ 
  cutler xx ecab xxx xxxx & 0.00 &  &  &  &  \\ 
  depression & 0.00 & 0.00 &  &  &  \\ 
  lillian cutler xx & 0.00 & 0.30 & 0.30 &  &  \\ 
  compensable factor of employment and &  & 0.01 & 0.01 &  &  \\ 
  results from an &  &  &  & 0.06 & 0.06 \\ 
  environment or &  &  &  & 0.01 & 0.01 \\ 
  an administrative or personnel &  &  &  & 0.01 & 0.01 \\ 
  requirement imposed &  &  &  & 0.01 & 0.01 \\ 
  allegations &  &  &  & 0.00 & 0.00 \\ 
   \hline
\end{tabular}
\caption[Summary of cases involving an ``emotional condition'' and causality]{Summary of cases involving an ``emotional condition'' and a discussion of causality.  Different columns correspond to the number of dropped phrases.  First column is only ``emotion'' and ``condition.''  Rest are adding more and more phrases.  Column B adds ``ecab,''  Column C ``depression,'' and Column D  ``xx'' (for illustration purposes). Column D includes many other terms.} 
\label{tab:emotion}
\end{table}

Column 1 of Table~\ref{tab:emotion} shows a first pass summary of those cases that both involve an emotional condition and revolve around issues of causality.
We see fairly general terms and some boilerplate language.
Here it is necessary to explore the raw text to discover the contexts for these phrases.
This is easily done using our package.
For example, one positively marked decision has:
\begin{quote}
\small
Not every injury or \textbf{illness that is} related to a claimant's employment is compensable. In the case of \textbf{Lillian Cutler}, the Board explained some distinctions as to the type of employment situations giving rise to a compensable emotional condition under FECA.

C.E., Docket No. 10-461 (issued November 23, 2010) (emphasis added).
\end{quote}
Another is:
\begin{quote}
\small
To establish that an emotional condition arose in the performance of duty, a claimant must submit the following: (1) medical evidence establishing that she has an emotional or psychiatric disorder; (2) \textbf{factual evidence identifying} employment factors or \textbf{incidents alleged to} have caused or contributed to the condition; and
(3) rationalized medical opinion evidence establishing that the emotional condition is \textbf{causally related} to the identified compensable employment factors.

T.G., 58 ECAB 189 (2006) (emphasis added).
\end{quote}

As an illustration of stability of CCS, consider the other columns of Table~\ref{tab:emotion}.
Each column corresponds to dropping more and more terms from consideration.
Note that the transition from the third to fourth column drops one of the case-references, ``Lillian Cutler,''  \emph{even though we did not explicitly drop those words and phrases.}  
CCS selected phrases are picked in the context of other phrases.  
Because we removed ``xx'' (indicating a case number), ``Lillian Cutler'' is no longer selected, along with other phrases including parts of this phrase and ``xx.''
Instead we obtain a cluster of phrases showcasing different aspects of these cases.
Dropping phrases can only affect a summary if those phrases are in the summary.
The final two columns are the same because none of the additional phrases were in the summary from column 4.  
Care must be taken to understand the complex dependencies between phrases.

Thus, in the context of ECAB decisions, the CCS tool provides phrases that flag boilerplate language and case citations. To some degree, these phrases appear to reflect precedent and common statements of law that characterize a given category of cases. 
While our results are exploratory, inexact, and not particularly revealing in and of themselves, they do suggest that a refined CCS tool might one day facilitate the development of automated case content analysis or aid the development of refined legal taxonomies.

\subsection{Discussion}
As the above studies illustrate, using these tools to understand text is a very different, and far less precise, activity than working to correctly classify text.
The common problems with machine-learning approaches (selecting methods, selecting tuning parameters, etc.) are only exacerbated by this uncertain area.
The researcher is left with many decisions to make, and only vague guidance on how to make them.

With our method, two such decisions are prominent: what method to use for selection of the regularization parameter $\regC$, and what method to use for selection of the feature-normalization parameter $q$.
We also need to determine how to remove domain-specific stop words.

\paragraph{Picking $\regC$.} 
For selecting $\regC$, we have several options, especially if we include picking a regularization parameter by optimizing predictive performance.
Which option to select is a difficult, especially since there is no easy metric of final quality if one's focus is not prediction.
The computational investigations shed some light on this problem.
Ideally one would use the maximum of the permutation approach and the rare-phrase pruning approach.
This will guarantee only finding a summary if one should be found, and also will discard rare phrases that do not speak of general trends across the positive documents.
The free test from the permutation-selected $C$ of whether the phrases as a whole are in fact significantly associated with the text is a real boon, in our view.
It moves towards presentation of results that are known to not be entirely noise.

That being said, future work on stability (where documents are perturbed to see how the selected phrases change, for example) is a must.
Furthermore, we acknowledge that the above examples suggest that the permutation-selected $C$ is  severe, more severe than from cross validation or similar.
This means we can lose human meaning as illustrated by the Methylene Chloride example: the richness of the summary was much greater with a slightly reduced $C$.
Relaxing regularization towards what would be achieved with prediction-oriented approaches  to achieve longer lists may be informative, but (other than improved prediction) this could undermine the guarantees provided by the above.
Perhaps work on testing individual phrases via false discovery rates could find a better balance.

Regardless, one should always compare the finally chosen $C$ to Table~\ref{tab:need_cut} to see to what degree it is discarding perfect predictors, and to what degree it would leave the remainder to be potentially picked up.
This provides a human interpretation of the impact of the regularization.

\paragraph{Picking $q$.}
Selecting $q$ is also admittedly difficult.
By design, it gives different views of the data, from the quite general to the very specific.  
We advocate for exploring a range of values as the best practice.
In the above case studies, for example, the full range of terms on the tables provided the most complex and rich story; perhaps pooling the lists and exploring these pooled lists would be one way forward.
We underscore that we view these tools as exploratory; the researcher can extract small snippits of text to see if they offer some clue towards a more thorough investigation.
This is similar in spirit to, for example, XGobi \cite{buja1996interactive}.

\paragraph{Stop words.}
Even though we avoid using stop-word lists as a first-pass approach, we still needed to generate specialized stop-word lists.
We see no way to have removed these words without human intervention as, from a prediction standpoint, the removed words are key indicators of a given labeling.
Unfortunately, selecting them occludes terms that could enhance human understanding.
We can see this with the calculated $C$ thresholds: as we add words to the ban list, the $C$ plummets because we are removing the words that are most correlated with the labeling.
Eventually we could reach a point where we have ``conditioned out'' all the connection of the labeling to text.
This is another potential avenue for exploring such data.
Overall, we advocate generating modest-length stop-word (ban) lists using substance-matter knowledge, coupled with rescaling, over using generic stop lists that allow milquetoast words through due to their being not obviously wrong.

\section{Conclusions}

We present a method  for comparing sets of documents that is simple, sparse, and fast.  
We argue that these qualities are important for text analysis, especially if it is to be used for surveillance or other exploratory tasks.
Here ``simple'' means the summaries cannot be too technical in nature.  
For example, the presence/absence of features is easier to interpret than regression weights.
We need sparsity as humans are lazy; the number of phrases in a summary must be few.
The faster summaries can be computed, the better. 
Otherwise exploratory analysis and discovery are bogged down.

We do not, however, argue that the results of these tools, or in fact any other text analysis tools that we know of, can be taken as ultimate proof of any particular substantive theme or meaning. Summaries can be quite suggestive, but researchers would need to investigate further to substantiate those suggestions.
Alternatively, secondary analyses such as blind scoring of the key phrases for sentiment could lead to traditional statistical conclusions. 
In these cases CCS should be viewed as a dimension reduction tool, providing a targeted, small number of informative features for a very complicated form of data.

On the technical side, we have effectively provided an implementation of Lasso-style regression where the full set of features are dynamically created and the loss is a squared hinge loss rather than a normal quadratic loss.
This work shows that implementing sparse regression with greedy coordinate offers a viable direction for summarization using phrases rather than words.  
Furthermore, the approach of dynamically building features shows promise for other customizations such as skipping or dropping words to automatically detect related phrases, collapsing them into single features.

Admittedly more work needs to be done to optimize this particular implementation to see exactly how fast it could be; currently the algorithm is sub-optimal because it does not fully fit currently selected features at each iteration.
That being said, compared to methods with a pre-computed design matrix, it is comparably fast, is more flexible in the $n$-grams considered, and allows for some trickiness such as having gaps in the key-phrases (i.e., wild-card words) and the enforcement of non-negative weights.
Additionally, the \verb|textreg| package is quite natural to use, allowing users to avoid calculating the phrase matrix, and instead works with raw text.
It also easily allows for customization of different rescaling schemes other than $L^2$.

Going beyond text analysis, these methods also hint at ways of incorporating many interaction terms among features in high-dimensional regression. 
Phrase features are simply interactions of nearby word features, and thus similar bounding methods may exist.  This is another area for future exploration.

\section*{Acknowledgements}
We would like to thank three anonymous reviewers and an AE for their detailed comments; the paper has been much improved by this feedback.
Also enormous thanks to Matt Taddy for his invaluable and speedy support with his \verb|textir| package used in our comparison studies.
This work builds on conversations and ideas discussed in the StatNews research group in UC Berkeley led by Bin Yu and Laurent El Ghaoui.  
In particular thanks to Garvesh Raskutti for his ideas on the impact of different rescaling choices.
The authors are very grateful for these opportunities and inspirations.
Also thanks to Kevin Wu for a portion of the code in the R package, and to Janet Ackerman for collaboration on an earlier incarnation of this project and initial data collection.

\section*{Appendix A}

Appendix A consists of supplemental tables showing alternate summaries of the case studies discussed above.
Table~\ref{tab:CO_full} compares Carbon Monoxide narratives to all the other narratives.  Table~\ref{tab:CO_stem} demonstrates a series of summaries on a stemmed corpus.

\begin{table}[ht]
\centering
\tiny
\begin{tabular}{lrrrrrrrrr}
  \hline
phrase & q=1.2 & q=1.5 & q=2 & q=3 & q=4 & $\#$ reports & $\#$ tagged & \% tagged & \% phrase \\ 
  \hline
were using a gasoline & 0.28 & 0.31 &  &  &  &   5 &   5 & 100.00 & 2.00 \\ 
  propane powered floor & 0.28 &  &  &  &  &   5 &   5 & 100.00 & 2.00 \\ 
  their blood & 0.28 & 0.52 & 0.45 & 0.64 & 0.65 &   5 &   5 & 100.00 & 2.00 \\ 
  hemoglobin & 0.15 &  &  &  &  &   5 &   5 & 100.00 & 2.00 \\ 
  in the cold room & 0.15 & 0.16 &  &  &  &   5 &   5 & 100.00 & 2.00 \\ 
  they were found &  & 0.34 & 0.15 &  &  &   4 &   4 & 100.00 & 2.00 \\ 
  powered forklifts &  & 0.31 & 0.18 &  &  &   9 &   8 & 89.00 & 3.00 \\ 
  gasoline powered &  & 0.30 & 0.47 & 0.56 & 0.53 &  66 &  30 & 45.00 & 12.00 \\ 
  propane powered &  & 0.28 & 0.40 & 0.53 & 0.53 &  52 &  28 & 54.00 & 12.00 \\ 
  overexposure &  & 0.18 & 0.34 & 0.53 & 0.58 &  37 &  18 & 49.00 & 7.00 \\ 
  exhaust fumes &  &  & 0.13 &  &  &   6 &   5 & 83.00 & 2.00 \\ 
  exhaust &  &  & 0.03 & 0.10 & 0.11 & 196 &  35 & 18.00 & 14.00 \\ 
  calculated &  &  &  & 0.18 &  &  12 &   7 & 58.00 & 3.00 \\ 
  generator was &  &  &  & 0.13 & 0.17 &  24 &  10 & 42.00 & 4.00 \\ 
  employees were treated &  &  &  & 0.10 & 0.13 &  40 &  12 & 30.00 & 5.00 \\ 
  evacuated &  &  &  & 0.07 & 0.10 & 104 &  18 & 17.00 & 7.00 \\ 
  the fire department &  &  &  & 0.06 &  & 241 &  21 & 9.00 & 9.00 \\ 
  ventilation &  &  &  & 0.05 & 0.07 & 203 &  32 & 16.00 & 13.00 \\ 
  were treated &  &  &  & 0.05 & 0.04 & 217 &  22 & 10.00 & 9.00 \\ 
  a propane &  &  &  & 0.05 & 0.05 & 128 &  26 & 20.00 & 11.00 \\ 
  powered &  &  &  & 0.03 & 0.03 & 740 &  85 & 11.00 & 35.00 \\ 
  blood &  &  &  & 0.03 & 0.04 & 355 &  30 & 8.00 & 12.00 \\ 
  were hospitalized &  &  &  & 0.02 & 0.03 & 778 &  42 & 5.00 & 17.00 \\ 
  found &  &  &  & 0.01 & 0.02 & 2885 &  68 & 2.00 & 28.00 \\ 
  employees &  &  &  & 0.00 & 0.01 & 5575 & 126 & 2.00 & 52.00 \\ 
  stratton &  &  &  &  & 0.27 &  12 &   7 & 58.00 & 3.00 \\ 
  headaches &  &  &  &  & 0.16 &  36 &  10 & 28.00 & 4.00 \\ 
  source of the &  &  &  &  & 0.09 &  60 &  12 & 20.00 & 5.00 \\ 
  a gasoline &  &  &  &  & 0.07 &  70 &  26 & 37.00 & 11.00 \\ 
  passed out &  &  &  &  & 0.06 & 108 &  11 & 10.00 & 5.00 \\ 
  enclosed &  &  &  &  & 0.05 & 180 &  18 & 10.00 & 7.00 \\ 
  fire department &  &  &  &  & 0.03 & 778 &  36 & 5.00 & 15.00 \\ 
  cold &  &  &  &  & 0.03 & 219 &  16 & 7.00 & 7.00 \\ 
  hours &  &  &  &  & 0.01 & 1277 &  33 & 3.00 & 14.00 \\ 
  room &  &  &  &  & 0.01 & 1580 &  42 & 3.00 & 17.00 \\ 
   \hline
\end{tabular}
\caption[Different Summaries of Carbon Monoxide vs. other cases]{Different Summaries of Carbon Monoxide (when comparing against all other cases)} 
\label{tab:CO_full}
\end{table}

\begin{table}[ht]
\tiny
\centering
\begin{tabular}{lrrrrrrrrrr}
  \hline
phrase & q=1.2 & q=1.5 & q=2 & q=3 & q=4 & $\#$ reports & $\#$ tagged & \% tagged & \% phrase \\ 
  \hline
poison+ at+ & 0.33 & 0.12 & 0.10 & 0.08 & 0.03 & 17 & 17 & 100 & 7 \\ 
  propan+ power+ & 0.12 & 0.38 & 0.55 & 0.56 & 0.49 & 34 & 28 & 82 & 12 \\ 
  gasolin+ power+ & 0.11 & 0.38 & 0.54 & 0.56 & 0.54 & 43 & 32 & 74 & 13 \\ 
  for fresh+ & 0.10 & 0.19 & 0.10 &  &  & 4 & 4 & 100 & 2 \\ 
  poison+ & 0.09 & 0.36 & 0.49 & 0.53 & 0.53 & 141 & 96 & 68 & 40 \\ 
  an+ X hour+ & 0.04 &  &  &  &  & 6 & 6 & 100 & 2 \\ 
  were+ use+ a+ gasolin+ & 0.04 &  &  &  &  & 5 & 5 & 100 & 2 \\ 
  was oper+ a+ propan+ & 0.03 &  &  &  &  & 6 & 6 & 100 & 2 \\ 
  the cold room+ & 0.03 & 0.02 &  &  &  & 7 & 7 & 100 & 3 \\ 
  concret+ saw+ &  & 0.08 & 0.21 & 0.10 &  & 10 & 9 & 90 & 4 \\ 
  XXX cubic+ &  & 0.06 &  &  &  & 4 & 4 & 100 & 2 \\ 
  the generat+ was &  & 0.06 & 0.09 &  &  & 10 & 8 & 80 & 3 \\ 
  forklift+ were+ &  & 0.01 & 0.05 &  &  & 8 & 7 & 88 & 3 \\ 
  averag+ &  &  & 0.13 & 0.16 & 0.10 & 17 & 11 & 65 & 5 \\ 
  headach+ dizzi+ &  &  & 0.06 &  &  & 6 & 5 & 83 & 2 \\ 
  intern+ combust+ &  &  & 0.05 &  &  & 8 & 6 & 75 & 2 \\ 
  the cold &  &  & 0.04 &  &  & 17 & 10 & 59 & 4 \\ 
  exhaust+ &  &  & 0.03 & 0.07 & 0.08 & 142 & 38 & 27 & 16 \\ 
  generat+ &  &  & 0.03 & 0.05 & 0.03 & 110 & 31 & 28 & 13 \\ 
  sourc+ of the &  &  & 0.02 & 0.10 & 0.08 & 33 & 13 & 39 & 5 \\ 
  X hour+ &  &  & 0.01 & 0.03 &  & 57 & 18 & 32 & 7 \\ 
  hour+ &  &  &  & 0.03 & 0.04 & 362 & 47 & 13 & 19 \\ 
  found+ &  &  &  & 0.02 & 0.03 & 561 & 69 & 12 & 28 \\ 
  a+ gasolin+ &  &  &  & 0.02 & 0.02 & 48 & 26 & 54 & 11 \\ 
  cold &  &  &  & 0.01 &  & 66 & 16 & 24 & 7 \\ 
  fire+ depart+ &  &  &  & 0.01 & 0.01 & 217 & 37 & 17 & 15 \\ 
  ventil+ &  &  &  & 0.01 & 0.01 & 173 & 42 & 24 & 17 \\ 
  power+ &  &  &  & 0.01 & 0.02 & 459 & 95 & 21 & 39 \\ 
  were+ treat+ &  &  &  & 0.01 & 0.01 & 112 & 22 & 20 & 9 \\ 
  were+ &  &  &  & 0.01 & 0.01 & 2778 & 168 & 6 & 69 \\ 
  employe+ were+ &  &  &  & 0.01 & 0.00 & 612 & 67 & 11 & 28 \\ 
  room+ &  &  &  & 0.00 & 0.00 & 475 & 43 & 9 & 18 \\ 
  employe+ are &  &  &  &  & 0.01 & 305 & 37 & 12 & 15 \\ 
   \hline
\end{tabular}
\caption{Different Summaries of Carbon Monoxide (using stemming, compared to chemical family)} 
\label{tab:CO_stem}
\end{table}

\section*{Appendix B: Derivations}
We here show three derivations used in the above work.
We first show how to obtain the bound on the gradient.
Second, we give an alternate formulation of the loss function which gives a different approach for finding the feature with the maximal gradient.
We then show how to obtain the minimal $\lambda^*$ to ensure perfect predictors are pruned.

\subsection*{Bound on the gradient}
The gradient for phrase $j$ is
\begin{align*}
  \frac{d}{d \beta_j }\mathcal{L}(\beta) &= \sum_{i=1}^n \xi'( m_i ) y_i \frac{c_{ij}}{z_j} + \regC  \frac{d}{d \beta_j} |\beta_j|  \\
  			&= \sum_{y_i = 1, c_{ij} > 0} \xi'( m_i ) \frac{c_{ij}}{z_j} + \sum_{y_i = -1, c_{ij} > 0} -\xi'( m_i ) \frac{c_{ij}}{z_j}  + \regC \frac{d}{d \beta_j} |\beta_j|
\end{align*}

Now consider all phrases $k$ with phrase $j$ as a prefix that are currently not in the model (i.e., which currently have $\beta_k = 0$), and maximize over the possible gradients (a similar argument gives a bound below for negative gradients).
For vectors $a, b$, let $a \preceq b$  denote a component-wise relationship of $a_i \leq b_i$ for $i = 1, \ldots, m$.
Then the set $\left\{ a : 0 \preceq a \preceq c_j\right\}$ contains all potential appearance patterns for a phrase with phrase $j$ as a prefix.
We do not wish to calculate what the actual phrases are, hence we optimize over this set of potential phrases.
This results in the optimization problem:
\begin{align*}
	U &= \max_{0 \preceq a \preceq c_j} \sum_{y_i = 1, c_{ij} > 0} \xi'( m_i ) \frac{a_i}{z_a} + \sum_{y_i = -1, c_{ij} > 0} \!\! -\xi'( m_i )  \frac{a_i}{z_a}  + \regC  \frac{d}{d \beta_j} |\beta_j| \\
	 &\leq  \max_{0 \preceq a \preceq c_j} \sum_{y_i = -1, c_{ij} > 0} \! \! -\xi'( m_i ) \frac{a_i}{z_a} - \regC =  \max_{0 \preceq a \preceq c_j} \frac{1}{z_a} \inner{w}{a} - \regC 
\end{align*}
with $z_a = |a|_q$ being the $L^q$ norm of $a$, $w$ being a $m$-vector of weights with $w_i = -\xi'( m_i )$, and $\inner{\cdot}{\cdot}$ being the inner product.

Because $\xi'(m_i)$ is everywhere nonpositive, the first term in the first line above is negative.
The second line follows because setting $a_i = 0$ for any document with $y_i = 1$ only increases the gradient as doing so will simultaneously drop negative terms and shrink $z_j$.  
The penalty term is negative because we are examining gradients at $0$; if we step $\epsilon$ in the negative direction (as indicated by the first term), the gradient will immediately be shrunk towards 0.
The $m_i$, which include the intercept $\mu$, are fixed constants, determined by the current location of our optimization path.
We are effectively maximizing over an inner product of $a$ and a vector of weights $w$ with $w_i = \xi'(m_i)$ for $i=1, \ldots, n$.
Overall, this bound is assessing the maximum possible utility of a hypothetical super phrase, which boils down to maximizing weights put on positive examples.

The normalization $z_a$ renders this problem difficult.  We can bound the optimization using the following relationship:
\[ |x|_2 |w|_2 \leq |x|_q |w|_r \]
for $q, r$ such that $1/q + 1/r = 1$.

This inequality gives
\begin{align*}
	U &=  \max_{0 \preceq a \preceq c_j} \frac{1}{|a|_q} |a|_2 |w|_2 \cos \theta - \regC \\
	&\leq \max_{0 \preceq a \preceq c_j}  \frac{1}{|a|_q} |a|_q |w|_r \cos \theta - \regC \\
	&= \max_{0 \preceq a \preceq c_j} |w|_r \cos \theta - \regC \\
	&\leq |w|_r  - \regC 
\end{align*}
with $\theta$ being the angle between $a$ and $w$.

Coupling this with a similar argument for minimization gives the overall bound.

\paragraph{A note on the Elastic Net.}
The Elastic Net \cite{Zou:2005wy} is where we penalize our loss function with (using Ifrim et al.'s notation)
\[ \regC R_a( \beta ) = \regC a  \sum_{j=1}^p \left| \beta_j \right|  +  \regC (1-a)  \left( \sum_{j=1}^p \left| \beta_j \right|^2 \right)^{1/2} .\]
This regularization tends to keep groups of correlated features rather than picking one; it can borrow from the stability of ridge regression.  It can be potentially useful when many small features have weak signals.  Setting $a=1$ corresponds to  $L^1$ regularization.

For our problem, the gradient search is just changed to a subtraction of $\regC a$ rather than $\regC$ for the at-zero potential new features.  The gradients calculated for features already in the model have an extra term of $\regC (1-a) 2 \beta_j$ due to the derivative of the second term above.

\subsection*{Alternate gradient formulation.}
By redefining  $\beta$, we can change the optimization problem to have a $L^q$-rescaling term in the penalty. 
This gives different bounds on the gradients for super-phrases based on a sub-phrase.
However, it also changes the gradients themselves, which would change the path of the optimization problem.
That is, different features will initially have the largest gradient.
Assuming true convergence, the final solution will be identical, however.

In particular, define $\tilde{\beta}_j = \beta_j / z_j$.  The loss term can then be reexpressed  as
\begin{align*}
 \mathcal{L}(\beta) &= \sum_{i=1}^n \xi( m_i ) + \regC \sum_{j=1}^p z_j |\tilde{\beta}_j| .
\end{align*}

The gradient for phrase $j$ is then
\begin{align*}
  \frac{d}{d \beta_j }\mathcal{L}(\beta) &= \sum_{i=1}^n \xi'( m_i ) y_i c_{ij} + \regC z_j \frac{d}{d \beta_j} |\beta_j|  \\
  			&= \sum_{y_i = 1, c_{ij} > 0} \xi'( m_i ) c_{ij} + \sum_{y_i = -1, c_{ij} > 0} \!\!\!-\xi'( m_i ) c_{ij} + \regC z_j \frac{d}{d \beta_j} |\beta_j|.
\end{align*}

Again consider all phrases with phrase $j$ as a prefix and maximize over the gradient, yielding the optimization problem:
\begin{align*}
	U &= \max_{0 \preceq a \preceq c_j} \sum_{y_i = 1, c_{ij} > 0} \xi'( m_i ) a_i + \sum_{y_i = -1, c_{ij} > 0} -\xi'( m_i ) a_i + \regC z_j \frac{d}{d \beta_j} |\beta_j| \\
	 &=  \max_{0 \preceq a \preceq c_j} \sum_{y_i = -1, c_{ij} > 0} \left|\xi'( m_i )\right| a_i - \regC \left( \sum_{i=1}^n a_i^p \right)^{1/p} .
\end{align*}

As before, the second line comes from noticing that setting $a_i = 0$ for any document with $y_i = 1$ will only increase the gradient due to dropping the negative terms and shrinking $z_j$.  
The penalty term is still negative because if we step $\epsilon$ in the negative direction from 0 (which is indicated by the first term), the gradient will immediately shrink towards 0.

If we consider only phrases that have at least $r$ occurrences in our corpus, then we can roughly bound with
\[  \max_{k : x_k \preceq x_j} \frac{d}{d \beta_k} L( \beta ) \leq \sum_{y_i = -1, c_{ij} > 0} \left| \xi'( m_i ) \right| c_j - \regC m^{1/p} . \]
This is from maximizing both terms separately.  For the first, we simply add maximum weight, without regard to the normalizing constant.
For the normalizing constant, given a total count of $r$ occurrences, the maximum $z_j$ would be putting singletons on each of $r$ documents, giving the $r^{1/p}$ total.

Similarly, bounding from below gives an overall bound of
\[ \left|  \max_{k : x_k \preceq x_j} \frac{d}{d \beta_k} L( \beta )  \right| \leq 0 \bigvee \max \left\{  \sum_{y_i = -1, c_{ij} > 0} \left| \xi'( m_i ) \right| c_{ij},  \sum_{y_i = 1, c_{ij} > 0} \left| \xi'( m_i )\right| c_{ij}   \right\} - \regC r^{1/p} \]

This bound appears to be less useful than the one presented in the main paper.  
Furthermore, not rescaling by $z_j$ tend to make more common phrases be selected first, as we are not rescaling the first term, allowing it to grow quite large.

\subsection*{Perfect predictors}
Take the count vector for a perfect predictor $c_j$.  It has $r$ $1$s and $m-r$ 0s.
For the regression, the count vector $c_j$ is $q$-rescaled, giving
\[ x_j = \frac{1}{z_j} c_j = \frac{1}{r^{1/q}} \qquad \mbox{as} \qquad z_j = \left( \sum_{i=1}^m |c_{ij}|^q \right)^{1/q} = r^{1/q} .\]

Assume feature $c_j$ has been set aside and we have optimized without it.  
We have $\hat{Y} = X\tilde{\beta}$ (with $\tilde{\beta}_j \equiv \beta_j / z_j$ except for the intercept) for our current set of predictions, and an overall predicted mean $\mu = \frac{1}{m}\sum \hat{Y}_i$.  
Now reintroduce feature $c_j$. 
Our loss function, when only considering feature $c_j$, is then
\begin{align*}
	 \ell(\beta_j) &= \sum_{i=1}^m \left(Y_i - \hat{Y}_i\right)^2 + \regC|\beta_j| \\
				&= \sum_{i=1}^m \left(Y_i - X_i \tilde{\beta} - x_{ij}\beta_j\right)^2 + \regC|\beta_j| \\
				&= \sum_{i : x_{ij > 0}} \left(1 - X_i \tilde{\beta} - \frac{\beta_j}{r^{1/q}}\right)^2 + \regC|\beta_j|
\end{align*}
where we have dropped those terms not dependent on $\beta_j$.
This is convex.  
Take the derivative and set equal to 0 to find the minimum:
\begin{align*}
	 \ell'(\beta) &= -\frac{2}{r^{1/q}} \sum_{i : x_{ij > 0}} \left(1 - X_i \tilde{\beta} - \frac{\beta_j}{r^{1/q}}\right) + \regC sgn(\beta) \\
	  	&= -\frac{2}{r^{1/q}} \sum_{i : x_{ij > 0}} \left(1-X_i \tilde{\beta} \right) + \frac{2}{r^{1/q}}\sum_{i : x_{ij > 0}} \frac{\beta_j}{r^{1/q}} + \regC sgn(\beta) \\
	  	&= -\frac{2}{r^{1/q}} \sum_{i : x_{ij > 0}} \left(1-X_i \tilde{\beta} \right) + 2\beta_j r^{1-2/q} + \regC sgn(\beta).
\end{align*}
The $\beta$s will not be negative, and hence we examine the positive case, allowing us to drop the $sgn()$ term.
Set $\ell'(\beta)$ equal to 0 and solve, giving
\[ \hat{\beta}_j = r^{1/q}\left( \frac{1}{r} \sum_{i : x_{ij > 0}} \left(1-X_i \tilde{\beta} \right) \right) - \frac{1}{2}r^{2/q-1} \regC . \]
The term in the outer parenthesis is the average prediction for the documents having the perfect predictor $c_j$.  

If, for a document $i$ with $c_{ik} = 1$, there are not really any other predictive features, then $\hat{Y}_i = X_i \tilde{\beta} \approx \mu$.  
If this is true for all of the documents predicted by $c_k$, then the above is then approximately
\[ \hat{\beta}_j \approx r^{1/q - 1} r (1-\mu) - \frac{1}{2}r^{2/q-1} \regC =  r^{1/q} (1-\mu) - \frac{1}{2}r^{2/q-1} \regC . \]
If some documents are predicted by other features included in the model, then the sum will be less and the necessary $\regC$ for pruning will be reduced.
Rearrange to obtain an approximate cut-off for $\regC$ to drop all perfect predictors that perfectly predict $r$ documents.

\paragraph{The $q=2$ case.}

For $q=2$ we have
\begin{align*}
  \hat{\beta}_j &= \sqrt{r}\frac{1}{r}\left( \sum_{i : x_{ij > 0}} \left(1-X_i \tilde{\beta} \right) \right) - \frac{1}{2} \regC  \\
   &\approx \sqrt{r} (1-\mu) - \frac{1}{2} \regC .
 \end{align*}
For the positive examples, $x_{ij} = 1/\sqrt{r}$, giving prediction of
\[ \hat{Y}_i \approx \mu + \frac{1}{\sqrt{r}} \left( \sqrt{r} (1-\mu) - \frac{1}{2} \regC \right) = 1 - \frac{1}{2\sqrt{r}}. \regC \]
The first term of  $\hat{\beta}_j$ takes our prediction perfectly to $+1$ and the second term shrinks the coefficient away from 1 by half of $\regC$.   
Predictions from predictors that predict for more documents will be shrunk less than those for fewer.
The raw coefficients will also be larger.

\section*{Appendix C: Using the textreg package}

Our text regression package, \verb|textreg| on CRAN, is a wrapper for an extensively modified version of the C++ code written by Georgiana Ifrim.
It is also designed to integrate with the \verb|tm| package, a commonly used R package for dealing with text.
Our package is fully documented, but it is research code, meaning gaps and errors are possible; the author would appreciate notification of anything that is out of order.
 
The primary method in this package is the regression call \verb|textreg()|. This method takes a corpus and a labeling vector and returns a \verb|textreg.result| object that contains the final regression result along with diagnostic information that can be of use.  The (somewhat edited) function heading along with default values is:
\begin{Verbatim}
textreg <- function(corpus, labeling, banned=NULL, 
          C = 1.0, Lq = 2,
          maxIter = 40,
          verbosity = 1,
          positive.only = FALSE, binary.features = FALSE, 
          no.regularization = FALSE,
          min.support = 1, min.pattern = 1, max.pattern = 100, gap = 0,
          convergence.threshold=0.0001 )
\end{Verbatim}
The main arguments to this method are listed below:
\begin{description}
\item[corpus] A vector of strings or a Corpus object built out of strings.
\item[labeling] A vector of +1/0/-1 values where 0 means drop from consideration.
\item[banned] A vector of unigrams (words) that should not be allowed in any summary phrase.
\item[C] The $\regC$ tuning parameter for regularization.
\item[Lq] The $q$ for the $L^q$-rescaling of terms.  10 or above is treated as infinity.
\item[maxIter]  The maximum number of iterations allowed before terminating even under no convergence.
\item[verbosity] 0 means silent.   Larger numbers mean more diagnostic printout.
\item[positive.only]   Only allow positive features (other than the intercept). Useful if there are few positive documents and many negative, baseline, documents.
\item[binary.features]   The feature vectors are 0-1 vectors indicating whether a phrase is in or not in any given document. This is compared to vectors of counts of how many times a phrases in a document. These feature vectors are $L^q$-rescaled regardless.
\item[no.regularization]   If TRUE then features will not be rescaled (which recovers the Ifrim et al. algorithm).
\item[min.support]  Phrases that do not appear this many times are not considered viable features. Increasing this number can greatly decrease the running time of the algorithm, but it will force the dropping of very rare phrases regardless of rescaling or regularization choice.
\item[min.pattern,max.pattern]  Minimum and maximum lengths for phrases that are considered.
\item[gap]  Number of words that can appear in a gap. A phrase can have multiple gaps  of this length.
\end{description}

The resulting \verb|textreg.result| object can be printed, plotted, and explored.  Try, in an R Console, typing \verb|rs| by itself or \verb|plot( rs )|.
 The method \verb|reformat.textreg.model(rs)| gives a nice table (see, e.g., Table~\ref{tab:meth_chl}) of summary statistics for the final phrases.
The side-by-side summary table such as Table~\ref{tab:CO} is made by passing a list of \verb|textreg.result| objects to \verb|make.list.table()|.
The method \verb|calc.loss( rs )| gives the final loss of a result \verb|rs| and \verb|predict( rs )| will return individual document-level predictions of the labeling.  
The method \verb|rule.to.matrix( rs )| gives back the $n \times r$ design matrix for the final selected $r$ phrases including intercept.

 To pick a tuning parameter one can use
\begin{Verbatim}
Cs <- find.threshold.C( corpus, labeling, ban.words, R=100 )
\end{Verbatim}
This method returns a $R+1$ length list of numbers. The first number is the choice of $\regC$ that will return a null model for the labeling given, and  the subsequent $R$ numbers constitute our found $\regC$ values that return a null model under a random permutation of the labeling (holding the zeros fixed).  
It takes the same parameters as \verb|textreg| except for \verb|maxIter| and \verb|C|. Be sure to use the same remaining values for both calls so that \verb|find.threshold.C| culminates with a $\regC$ corresponding to the correct model family.

For exploring text, \verb|sample.fragments( phrase, labeling, corpus )| is useful.  See also \verb|grab.fragments()|.  To profile specific phrases, possibly even phrases not in the results, use \verb|make.phrase.count.table()|.

One can make cluster plot of how phrases relate with \verb|cluster.phrases( rs )|, or make matrices of co-occurrence of phrases using \verb|make.phrase.correlation.chart( rs )|. 
 
All of the above, and a bit more, is demonstrated and more fully explained in the vignette ``Bathtub Demo,'' that comes with the package.
Please read through it for further discussion and ideas.
 
\bibliographystyle{unsrt}
\bibliography{all_references}



\end{document}